\documentclass[lettersize,journal]{IEEEtran}
\usepackage{amsmath,amsfonts}
\usepackage{algorithmic}
\usepackage{algorithm}
\usepackage{array}
\usepackage{textcomp}
\usepackage{stfloats}
\usepackage{url}
\usepackage{verbatim}
\usepackage{graphicx}
\usepackage{cite}
\usepackage{xcolor}
\usepackage{booktabs}
\usepackage{multirow}
\usepackage{caption}
\usepackage{subcaption}
\usepackage[hidelinks]{hyperref}
\usepackage{orcidlink}

\begin{document}

\title{Spotter: Efficient Urban Visual Localization via Geo-Referenced Facade Landmarks in GPS-Degraded Environments}

\author{Antoni~Valls~\orcidlink{0009-0004-5511-2795}, Jordi~Sanchez-Riera~\orcidlink{0000-0002-4803-5742}%
\thanks{*Both authors are with the Institut de Rob\`otica i Inform\`atica Industrial,
CSIC-UPC, Llorens i Artigas 4-6, 08028 Barcelona, Spain
(e-mail: avallsc@iri.upc.edu; jsanchez@iri.upc.edu).}
}

\markboth{}
\markboth{}%


\maketitle

\begin{abstract}

Accurate visual localization on robotic and wearable platforms remains challenging in dense urban environments. Existing methodologies typically rely on GPS for absolute positioning, yet GPS signals frequently degrade in urban canyons due to multipath propagation. Consequently, standard solutions like visual odometry suffer from unmitigated drift over time, while map-matching techniques struggle to acquire the reliable GPS priors they need, on top of being too computationally heavy for real-time edge execution. To address these limitations, we propose Spotter, a robuts and real-time visual localization framework that uses building facades as a reliable source of global geo-reference, while retaining the capability to integrate GPS signals when available. In an offline stage, Spotter processes Google Street View panoramas by semantically segmenting facades and pairing multi-view stereo depth with cartographic data to build a compact metric database. At runtime, query images are matched via a cascaded retrieval and geometric verification pipeline to recover fine-grained global camera localization. We benchmark Spotter on a newly collected dataset of pedestrian sequences acquired with wearable smart glasses across several districts of Barcelona. Experimental results show that Spotter outperforms odometry-based baselines and achieves localization accuracy comparable to state-of-the-art map-based methods while operating at significantly higher frame rates.

\end{abstract}


\section{INTRODUCTION}

\begin{figure}[h!]
    \centering
        \includegraphics[width=\linewidth]{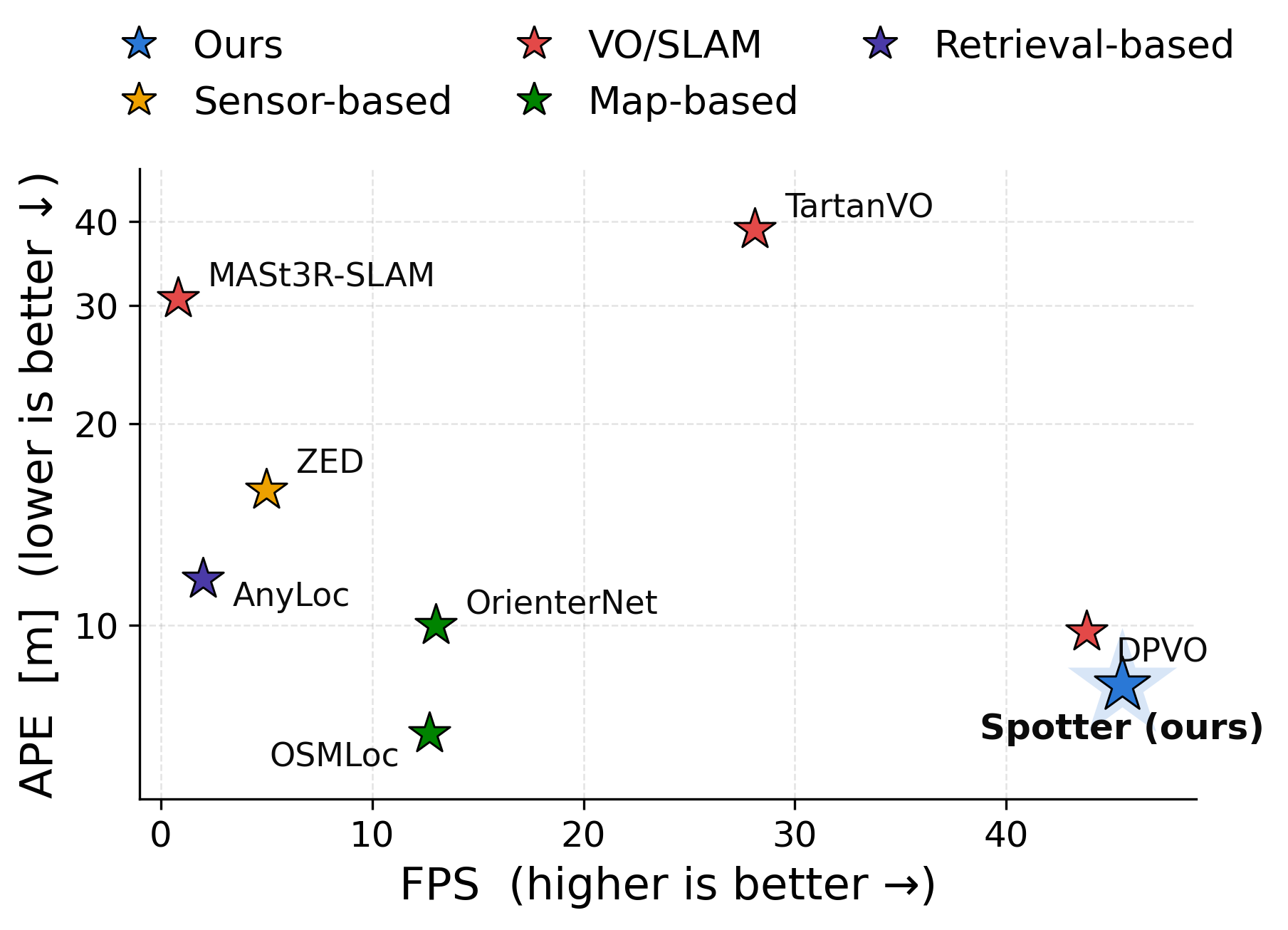}
    \caption{Accuracy--speed trade-off of different localization paradigms on our urban benchmark dataset. The $y$-axis reports the average position error, while the $x$-axis shows the inference speed in frames per second (FPS). Spotter achieves competitive localization accuracy while operating at significantly higher frame rates than existing map-based approaches. In addition, Spotter is robust to GPS degradation (see Section~\ref{sec:gps_ablation}), whereas the performance of GPS-assisted methods deteriorates as GPS uncertainty increases. Note that the reported accuracy of DPVO is obtained after global alignment to the ground-truth trajectory, which is not available during online operation. Where applicable, methods are shown in their GPS-aided configuration; see Table~\ref{tab:overall} for the corresponding values.}
    \label{fig:paradigms}
\end{figure}

\IEEEPARstart{R}{obot} navigation relies on precise visual localization to enable motion planning and seamless interaction with the surrounding environment. In urban settings, localization remains particularly challenging due to dynamic scenes, frequent occlusions, and the scale of the environment. These difficulties are most acute for lightweight robotic platforms and wearable devices with constrained computational resources. Whereas autonomous vehicles~\cite{shangg,BADUE2021113816} can afford a rich sensor suite---high-precision Global Positioning System (GPS) receivers, inertial sensors, LiDAR, and detailed high-definition maps---lightweight platforms~\cite{assist,Zhang,moisan2025mapin} such as smart glasses and smartphones operate under far stricter limits on sensing, computation, and energy. 

To operate within these strict resource limits, most practical navigation systems pivot toward Visual Odometry (VO)~\cite{dpvo,qiu2025mac} or Visual Simultaneous Localization and Mapping (SLAM)~\cite{orbslam3,driftfree,murai2024_mast3rslam}, which operate efficiently on embedded and wearable platforms through incremental feature-based motion estimation. However, this incremental formulation inevitably accumulates localization drift over time. The standard mechanisms for drift correction are poorly suited to autonomous navigation, particularly in pedestrian scenarios. Loop closures are unlikely to occur along the short, exploratory, and non-repetitive trajectories typical of pedestrian motion; GPS-based correction can be unreliable in dense urban environments due to signal blockage and multipath effects; and digital maps may be incomplete or inaccurate. Consequently, although these methods are computationally efficient and lightweight, they remain prone to unbounded localization drift in the absence of a reliable global reference, making accurate recovery of the global trajectory increasingly difficult.

To provide such a global reference, recent approaches have explored map-based image localization, estimating the camera pose independently at individual observations rather than incrementally over a complete trajectory~\cite{sarlin2023snap,sarlin2023orienternet,liao2024osmloc}. These methods typically project the query image into a bird's-eye-view representation and match it against rasterized OpenStreetMap (OSM) tiles, enabling pose estimation from lightweight cartographic maps. However, matching the query against a large set of potential candidate locations requires substantial computation, which can increase inference time. Moreover, these methods also rely on a GPS prior to constrain the search space and select candidate map tiles for matching, which can introduce GPS errors of tens of meters~\cite{wen2021graphgnss}, potentially expanding or misdirecting the search region. Consequently, localization performance can become sensitive to GPS inaccuracies, while the need to evaluate a larger set of candidate locations can further limit their suitability for real-time trajectory estimation.

A related approach formulates visual localization as an image retrieval problem. These approaches use pre-collected databases of geo-referenced images, typically obtained from Google Street View (GSV) or similar sources, and retrieve the most visually similar database image using global image descriptors~\cite{netvlad,monocularli}. Unlike incremental odometry, image retrieval can provide an absolute geographic reference without accumulating trajectory drift. However, this comes at the cost of storing and searching large collections of images, resulting in substantial memory and computational requirements. Furthermore, the retrieval process is not explicitly optimized for efficient large-scale localization, making its computational cost increasingly significant as geographic coverage and database size grow.

These limitations reveal a fundamental trade-off between localization accuracy, computational efficiency, and memory requirements, as illustrated in Figure~\ref{fig:paradigms}. Odometry and SLAM provide lightweight real-time operation but suffer from accumulated drift, whereas map-based and image-retrieval approaches provide global references at the cost of increased computational requirements and, in some cases, dependence on accurate GPS priors. Urban environments, however, contain a particularly valuable source of persistent visual information: building facades. Unlike transient objects such as pedestrians and vehicles, building facades are relatively stable, spatially distributed, and naturally associated with geographic coordinates. This makes them well suited to serve as visual landmarks for global localization, including in GPS-degraded or GPS-denied environments.

To exploit this property and address the limitations of existing localization approaches, we introduce \textbf{Spotter}, a global visual localization framework that leverages building facades as stable, geo-referenced visual landmarks. Spotter constructs a compact representation that combines discriminative facade features with metric world coordinates, enabling efficient image-based localization at city scale. Its design bridges the global consistency of map-based localization and the efficiency of feature-based visual matching, while avoiding the need for dense 3D reconstruction or continuous trajectory integration.

Spotter constructs its offline database from GSV panoramas by first semantically segmenting building facades and then extracting robust visual features associated with metric world coordinates using cartographic information and multi-view stereo depth estimation. At runtime, features extracted from a street-view query image are processed through a cascaded retrieval and geometric verification pipeline to efficiently identify corresponding facade landmarks and estimate the camera's global pose. GPS can optionally be used as a coarse spatial prior to restrict the search area, but it is not required for pose estimation, which is determined entirely from visual correspondences. Consequently, Spotter can operate with degraded GPS measurements or in a fully GPS-free configuration, providing fine-grained, drift-free localization at real-time frame rates within the computational and memory constraints of wearable platforms.

Our contributions can be summarized as:
\begin{itemize}
    \item A scalable offline pipeline that converts Google Street View (GSV) panoramas into compact, geo-referenced facade maps for efficient global pose estimation, eliminating the need for dense, computationally heavy Structure-from-Motion (SfM) reconstructions.
    \item A robust, real-time, and modular visual localization framework for wearable urban navigation that integrates seamlessly into existing navigation pipelines and supports both GPS-aided and GPS-free operation with minimal computational overhead.
    \item A new benchmark dataset for pedestrian visual localization, acquired using a wearable smart-glasses platform and comprising 20 challenging trajectories across diverse urban environments in several districts of Barcelona.
\end{itemize}


We conduct a comprehensive experimental evaluation on a pedestrian dataset collected using wearable smart glasses. In particular, we compare Spotter against representative VO/SLAM, map-matching, and retrieval baselines. We further evaluate GPS-free operation and robustness to increasing levels of synthetic GPS noise. The results demonstrate that Spotter achieves competitive or superior localization accuracy while providing substantially higher runtime efficiency and maintaining stable performance under degraded or absent GPS signals.

\section{RELATED WORK}

\subsection{VO/SLAM Localization}

VO and SLAM aim to estimate camera motion sequentially matching features across consecutive image frames. Unlike SfM approaches~\cite{sfm,sfmr,lynen,liu,Toft_2018_ECCV} that require large memory to store the reconstructed 3D dense point clouds, these methods execute efficiently with a minimal memory footprint. However, their primary limitation is the accumulation of drift over time, which eventually leads to trajectory degradation and loss of tracking. Recent advances have focused on enhancing feature-matching robustness, as seen in ORB-SLAM3~\cite{orbslam3} and MAC-VO~\cite{qiu2025mac}, aligning local SLAM point clouds against city-scale 3D meshes via Iterative Closest Point (ICP)~\cite{driftfree}, or utilizing dense optical flow across image sequences as in DPVO~\cite{dpvo}. Alternatively, multimodal approaches integrate complementary sensory inputs alongside visual features, IMUs, and GPS receivers~\cite{Cai2019MobileRL}. Most recently, 3D Gaussian Splatting has been explored as a scene representation for downstream mapping and pose estimation~\cite{sidorov2025gs, kerbl3Dgaussians}. Despite these advancements, incremental frameworks remain fundamentally vulnerable to unbounded drift—particularly when GPS signals are degraded—rendering long-range trajectory correction challenging without reliable global anchors.

\subsection{Map-based Localization}

Digital cartography provides a rich global cue that remains largely unexploited by standard VO and SLAM pipelines. Cartographic representations offer a key structural advantage: storing vector geometries requires an extremely compact memory footprint compared to 3D point clouds. To leverage this, recent data-driven frameworks directly match street-level images against learned cartographic feature maps. Specifically, OrienterNet~\cite{sarlin2023orienternet} transforms query images into bird's-eye-view (BEV) features to match against rasterized OSM tiles, achieving fine-grained 3-DoF pose recovery and leveraging temporal probability fusion for improved recall. OSMLoc~\cite{liao2024osmloc} enhances cross-city generalization by combining OSM semantic embeddings with geometry-guided depth estimation, while SNAP~\cite{sarlin2023snap} aggregates overhead and ground-level imagery into neural map representations for global-scale retrieval. Nevertheless, while these frameworks operate on lightweight maps and yield precise localization under accurate GPS priors, dense feature matching against 2D cartographic grids is computationally expensive. When GPS signals are degraded or unavailable, the required search radius expands dramatically, severely slowing down execution and limiting real-time deployment on wearable hardware.
 
\subsection{Retrieval-based Localization}

Rather than relying on learned vector map features, an alternative paradigm leverages digital cartographic maps to organize geo-referenced database imagery, such as GSV panoramas, for image retrieval. Classic Visual Place Recognition (VPR) methods like NetVLAD~\cite{netvlad, patch} represent scenes using sets of geo-tagged images, estimating global pose by identifying the nearest visual neighbor via global image descriptors. Foundation-model VPR approaches such as AnyLoc~\cite{keetha2023anyloc} extend this unconstrained retrieval paradigm, approximating the query pose directly using the database coordinates of the top-retrieved GSV panorama without requiring positional priors or downstream pose refinement. To achieve higher metric accuracy, methods such as Li et al.~\cite{monocularli} perform block-wise semantic segmentation on query images to match facade regions against GSV panoramas, subsequently refining the 6-DoF pose via bundle adjustment anchored to reference GSV coordinates. Nevertheless, storing dense image collections for wide geographic regions imposes significant memory overhead, while global query matching remains computationally expensive and often reliant on coarse GPS priors to bound the search domain.

\section{Spotter}

\begin{figure*}[t!]
    \centering
    \includegraphics[width=1\linewidth]{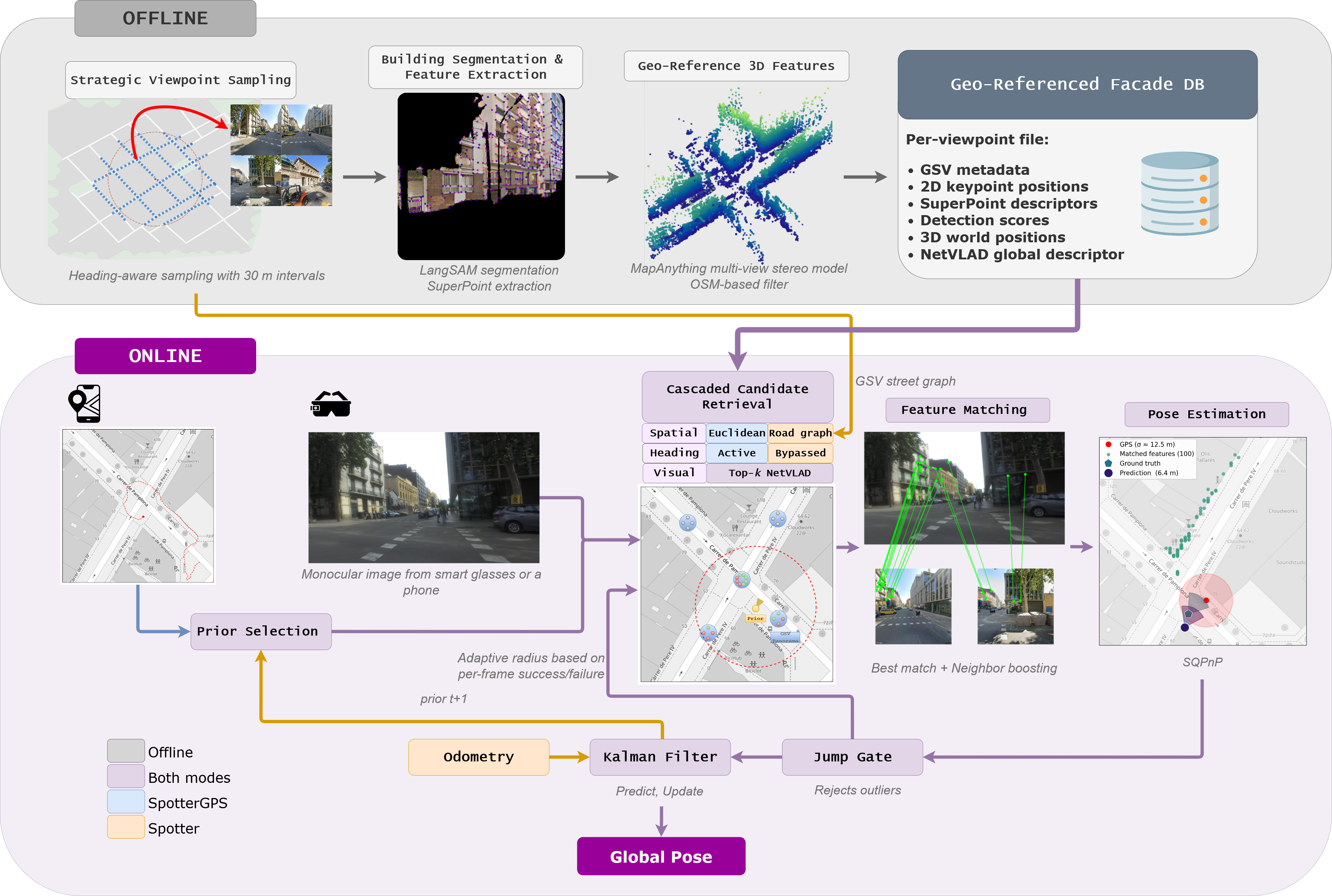}
    \caption{Overview of Spotter. \emph{Top}: the offline phase segments GSV panoramas, extracts keypoints, and lifts them to 3D world coordinates, building a compact per-viewpoint feature database and a street graph.
    \emph{Bottom}: at runtime a query image and an optional GPS signal drive candidate retrieval, feature matching, and pose estimation, with a Kalman
    filter smoothing the result. Connector and box color indicate which operating mode uses each path (see legend); the two modes share the same visual matching core and differ only in the spatial prior, the heading
    filter, and the propagation of the prior across frames.}
    \label{fig:pipeline}
\end{figure*}

Given a street-level query image $\mathbf{I}_t$, we aim to estimate the camera's 3-DoF global pose $\boldsymbol{\xi}_t = (\mathbf{g}_t, \theta_t) \in \mathbb{R}^2 \times (-\pi, \pi]$, where $\mathbf{g}_t = (e_t, n_t)$ denotes the UTM metric coordinates (easting, northing) and $\theta_t$ is the global heading. To achieve this, we design a two-stage framework (Figure~\ref{fig:pipeline}): an offline phase (Section~\ref{sec:offline}) that constructs a lightweight database $\mathcal{D}$ of 3D geo-referenced visual landmarks, followed by an online phase (Section~\ref{sec:online}). The online localization engine establishes 2D-3D feature correspondences and solves for the full 6-DoF camera pose $\mathbf{T}_t \in \mathrm{SE}(3)$ via Perspective-$n$-Point (PnP), from which $\boldsymbol{\xi}_t$ is analytically derived.

\subsection{Offline Geo-Referenced Database}
\label{sec:offline}

The construction of the geo-referenced database $\mathcal{D}$ follows a three-stage pipeline. First, a strategic viewpoint sampling strategy extracts panoramic imagery across a designated cartographic region. Second, building facades are semantically segmented to constrain the extraction of local and global visual features to persistent structural elements. Third, local 2D image features are projected into 3D metric world coordinates and geometrically verified against cartographic boundaries.

\subsubsection{Strategic viewpoint sampling} We define a geographic search space $\mathcal{R}(c_0, r)$ centered at geodetic coordinate $c_0 = (\text{lat}_0, \text{lon}_0)$ with radius $r$, and extract the underlying OSM road graph $\mathcal{G} = (\mathcal{N}, \mathcal{E})$. Along each road edge $e$ in $\mathcal{E}$, points are sampled at uniform metric intervals $\delta_s = 30\,\text{m}$, yielding a set of sample coordinates $S$. 

For each point $\mathbf{p}_k \in S$, GSV perspective crops are retrieved relative to the local road heading $\theta_k$. Specifically, we extract $N = 4$ views at $90^{\circ}$ orthogonal steps for straight segments $(\mathbf{p}_k \in S_{\text{segment}})$, and $N = 8$ views at $45^{\circ}$ directional steps for intersections $(\mathbf{p}_k \in S_\text{intersection})$. If the returned panorama position $\mathbf{p}_k^{GSV}$ differs from the target sample point $\mathbf{p}_k$, the database coordinate is updated to match the true GSV metadata anchor $\mathbf{p}_k^{GSV}$.

\subsubsection{Building segmentation and feature extraction} To eliminate dynamic visual clutter (e.g., vehicles and vegetation), each retrieved GSV image $\mathbf{I}_i$ at sampling location $\mathbf{p_k}$ is semantically segmented using LangSAM~\cite{medeiros2023langsam} to yield a binary building facade mask. Within the segmented facade region, we extract visual building landmarks using SuperPoint~\cite{superpoint}, represented as keypoint-descriptor pairs $\{(\mathbf{x}_{ij}, \mathbf{d}_{ij})\}_{j=1}^{K_i}$, where $K_i$ denotes the number of keypoints extracted from image $\mathbf{I}_i$ (capped at a maximum of $10{,}000$), $\mathbf{x}_{ij} \in \mathbb{R}^2$ denotes 2D image coordinates and $\mathbf{d}_{ij} \in \mathbb{R}^{256}$ is the corresponding local feature descriptor. In parallel, a global image descriptor $\mathbf{f}_{i} \in \mathbb{R}^{4096}$ is extracted via NetVLAD~\cite{netvlad}. The global vector $\mathbf{f}_{i}$, local features, and camera metadata $\mathcal{M}_{i}$ are indexed to enable coarse-to-fine visual retrieval during online execution.

\subsubsection{Geo-reference 3D features} 
\begin{figure}[t]
    \centering
    \includegraphics[width=1\linewidth]{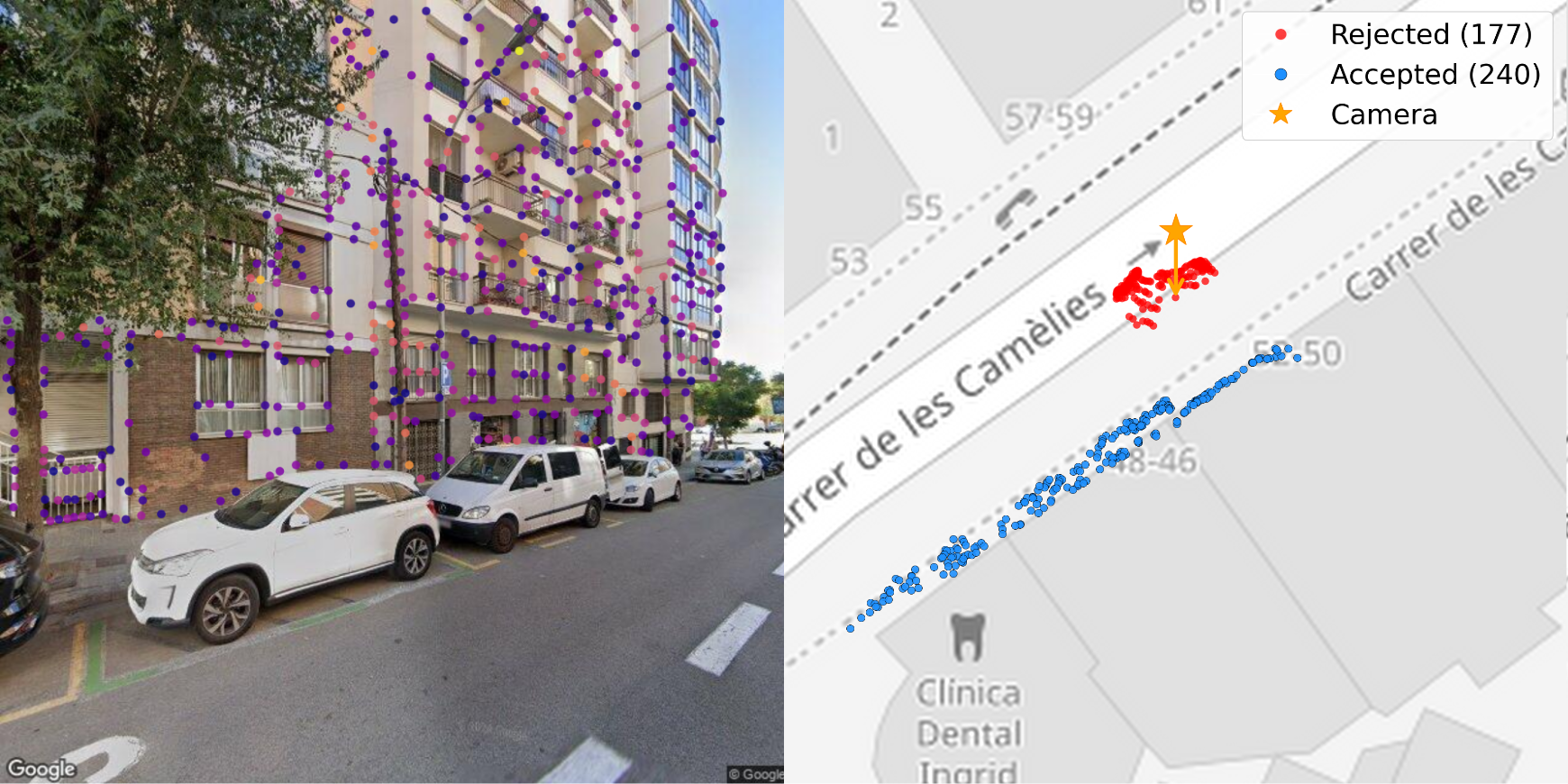}
    \caption{Offline database construction. \emph{Left}: SuperPoint keypoints extracted within the LangSAM building-facade mask on a GSV image. \emph{Right}: the same keypoints projected onto an OSM tile with their estimated 3D world positions; points discarded by the OSM facade-distance filter are shown in red.}
    \label{fig:offline}
\end{figure}

Each 2D keypoint $\mathbf{x}_{ij} \in \mathbb{R}^{2}$ is projected into a 3D cartographic landmark $\mathbf{X}_{ij} \in \mathbb{R}^3$ using per-pixel metric depth and an associated confidence score $s_{ij}$ predicted by a neural multi-view 3D estimator~\cite{mapanything}. To minimize database storage overhead and eliminate geometrically spurious estimates, candidate 3D landmarks undergo an OSM-guided spatial filtering stage illustrated in Figure~\ref{fig:offline}. Specifically, points whose 2D planar ground projections exceed a distance threshold $\tau_{\text{dist}}$ from actual building facade boundaries retrieved from OSM are pruned. This spatial verification step guarantees that only high-confidence facade landmarks are retained in the database.

Finally, the resulting geo-referenced database $\mathcal{D}$ consists of viewpoint tuples $\mathcal{V}_i = \left( \mathbf{f}_i, \mathcal{M}_i, \{(\mathbf{x}_{ij}, \mathbf{d}_{ij}, \mathbf{X}_{ij}, s_{ij})\}_{j=1}^{K_i} \right)$ collected across all sampled road graph locations $\mathbf{p}_k \in S$.

For a typical urban area of $A = 1\,\text{km}^2$, the generated database comprises approximately $V \approx 7{,}000$ viewpoint entries. With an average storage size of $\sim 100\,\text{KB}$ per viewpoint, the entire regional database occupies roughly $700\,\text{MB}$, maintaining a compact memory footprint suitable for edge deployment.

\subsection{Online Localization Engine}
\label{sec:online}

At runtime, the system localizes a temporally ordered sequence of query images indexed by the frame index $t = 1, 2, \dots$. At each step $t$, the model receives a query image $\mathbf{I}_t$ alongside a positional prior $\boldsymbol{\pi}_t$---an estimate of the camera's location used to seed the candidate search---and returns an estimated global camera pose $\boldsymbol{\xi}_t$. The architecture supports two distinct execution modes that differ in how $\boldsymbol{\pi}_t$ is obtained: \textit{SpotterGPS}, which uses the incoming GPS reading at frame $t$ as the prior (with estimated uncertainty $\sigma_t^{\text{GPS}}$); and \textit{Spotter}, which uses the previous Kalman filter (KF) state $\boldsymbol{\pi}_t = \hat{\mathbf{g}}_{t-1}^+$ as the prior for the candidate search region. 
Under both modes, each query image is localized against the reference database $\mathcal{D}$ through a four-stage pipeline: (1)~cascaded candidate retrieval, (2)~feature matching and pose estimation, (3)~Kalman filter position smoother, and (4)~Operating modes and gating mechanism.

\subsubsection{Cascaded candidate retrieval}
In a similar manner to the geo-referenced database construction, $K_t = 1{,}024$ SuperPoint keypoint-descriptor pairs $\{(\mathbf{x}_{tj}, \mathbf{d}_{tj})\}_{j=1}^{K_t}$ and a global NetVLAD descriptor $\mathbf{f}_t$ are extracted from the query image for downstream candidate matching. The pool of reference viewpoints $\mathcal{V}_i$ is then refined through three successive filtering stages to prune candidate space prior to computationally intensive feature matching. First, a \emph{spatial filter} retains only reference viewpoints located within a search radius $r_t$ centered on the current positional prior $\boldsymbol{\pi}_t$. 

The search radius $r_t$ is defined as:

\begin{equation}
    r_t = \begin{cases} 15.0 \cdot \lambda_t + \sigma_t^{\text{GPS}} & \text{SpotterGPS,} \\ 38.0 \cdot \lambda_t & \text{Spotter,} \end{cases}
    \label{eq:search_radius}
\end{equation}
where $\sigma_t^{\text{GPS}}$ is the estimated GPS uncertainty of the current prior and $\lambda_t \in [0.5, 2.0]$ is an adaptive scale factor initialized at $\lambda_1 = 1.0$. The scale factor adapts online:
\begin{equation}
    \lambda_{t+1} = \begin{cases} \max(0.5,\; \lambda_t \times 0.85) & \text{if frame } t \text{ succeeds,} \\ \min(2.0,\; \lambda_t \times 1.30) & \text{if frame } t \text{ fails,} \end{cases}
    \label{eq:radius_scale}
\end{equation}

shrinking the window after each success and widening it after each failure. The asymmetric factors ($\times 0.85$ vs.\ $\times 1.30$) cause the radius to reach its floor after approximately five consecutive successes and its ceiling after three consecutive failures. In SpotterGPS the spatial filter is a Euclidean radius filter, while in Spotter it is replaced by the road-graph-constrained filter of Section~\ref{sec:sequential}.

A \emph{heading filter} then discards references whose capture heading deviates from the estimated viewing direction by more than $120^{\circ}$; the heading is estimated from the GPS displacement history over a sliding window of roughly $6$\,s and requires at least $2$\,m of cumulative displacement before it is trusted, so this filter is active only in SpotterGPS and is bypassed in Spotter. Finally, a \emph{visual filter} keeps the top-$K_\text{filter}$ references by cosine similarity between the query and reference NetVLAD descriptors, reducing the candidate set to the most appearance-compatible viewpoints.

\subsubsection{Feature matching and pose estimation}
Candidate viewpoints $\mathcal{V}_i$ are subsequently processed in batches, sorted in descending order of visual similarity. For each candidate viewpoint, visual features are matched against the query image using LightGlue~\cite{lindenberger2023lightglue}. To remove false matches, geometric verification is performed per candidate by estimating a homography via MAGSAC~\cite{magsac}, filtering out correspondence outliers. Processing stops early once enough strongly-matched candidates have been found. The matched candidates are scored by a combined metric of inlier count and average match confidence, boosted by spatial-neighbor agreement; the top-scoring candidate is selected, and its 2D--3D correspondences are aggregated with those of up to eight spatially neighboring viewpoints that also matched the query. The final pose is recovered by applying the SQPnP solver~\cite{sqpnp} inside RANSAC~\cite{ransac}.

\subsubsection{Kalman filter position smoother}
A 2D Kalman filter tracks the estimated camera position $\mathbf{\hat{g}} = [e,\, n]^\top$ (easting, northing in UTM) with covariance $P \in \mathbb{R}^{2\times2}$ across frames. The filter is initialized on the first accepted localization with
\begin{equation}
    \mathbf{\hat{g}}_0 = [e_0,\, n_0]^\top, \qquad P_0 = \sigma_{p_0}^2\, I,
\end{equation}
where $\sigma_{p_0} = 20\,\text{m}$ for both SpotterGPS and the Spotter cold-start.

\paragraph{Predict} The prediction model differs between modes. In SpotterGPS, the state evolves as an undirected random walk:
\begin{equation}
    \mathbf{\hat{g}}^- = \mathbf{\hat{g}}, \qquad P^- = P + \sigma_q^2\, I,
\end{equation}
where $\sigma_q = v_{\max}\,\Delta t$, $v_{\max} = 2.5\,\text{m/s}$ is a conservative walking-speed bound, and $\Delta t$ is the inter-frame time interval. When a reliable heading $\theta^\text{GPS}$ is available from GPS displacement history, a directed variant is used instead, where $v^{\text{GPS}}$ is the GPS-measured speed capped at $v_{\max}$:
\begin{equation}
    \mathbf{\hat{g}}^- = \mathbf{\hat{g}} + \begin{bmatrix} v^{\text{GPS}}\,\Delta t\,\sin\theta^\text{GPS} \\ v^{\text{GPS}}\,\Delta t\,\cos\theta^\text{GPS} \end{bmatrix}, \qquad P^- = P + \sigma_q^2\, I.
\end{equation}
In Spotter, the state is propagated using the inter-frame displacement $[\Delta {x}^\text{odo}, \Delta y^\text{odo}]^\top$ provided by an external odometry source, aligned to the UTM frame via a rotation matrix $R_{\text{odo}}$ estimated online by Procrustes analysis:
\begin{equation}
    \mathbf{\hat{g}}^- = \mathbf{\hat{g}} + R_{\text{odo}}\begin{bmatrix}\Delta x^\text{odo} \\ \Delta y^\text{odo}\end{bmatrix}, \qquad P^- = P + \sigma_q^2\, I,
\end{equation}
with $\sigma_q = 0.5$\,m (trusted odometry), or falling back to $\sigma_q = v_{\max}\sqrt{\Delta t_{\text{succ}} \cdot \Delta t}$ when odometry is unavailable, where $\Delta t_{\text{succ}}$ is the elapsed time since the last successful localization. This wider process noise reflects growing positional uncertainty across longer localization gaps. The Procrustes alignment is fit once via singular value decomposition (SVD) as soon as at least five PnP--odometry pairs span a minimum cumulative motion of $5$\,m, and is then frozen for the remainder of the sequence. Before this alignment is established, all Spotter predictions rely on the undirected random-walk fallback, making the system most vulnerable to early localization failures before sufficient trajectory has been observed.

\paragraph{Update} Each accepted PnP estimate $\mathbf{z} = [e_{\text{pred}},\, n_{\text{pred}}]^\top$ is fused as a direct position observation:
\begin{gather}
    K = P^-\!\left(P^- + \sigma_r^2\, I\right)^{-1}, \qquad  P^+ = (I - K)\,P^-, \notag \\[4pt] \mathbf{\hat{x}}^+ = \mathbf{\hat{g}}^- + K(\mathbf{z} - \mathbf{\hat{g}}^-)
   .
\end{gather}
where the measurement noise is derived from PnP quality:
\begin{equation}
    \sigma_r = \mathrm{clip}\!\left(\frac{3.0 \times \mathrm{RMSE}_{\mathrm{px}}}{\sqrt{N_{\mathrm{inliers}}}},\; 2,\; 50\right)\,\text{m}.
\end{equation}

High reprojection error or few inliers produce a large $\sigma_r$, causing the filter to down-weight the measurement and lean on the prediction. 

 
\subsubsection{Operating modes and jump gate}
\label{sec:sequential}


In \emph{SpotterGPS}, the prior $\boldsymbol{\pi}_t$ for frame $t$ is set directly to the current GPS reading. Each frame is therefore anchored independently to a fresh GPS measurement, so a failed prediction on frame $t$ has no effect on frame $t{+}1$. The heading filter is active, using the GPS displacement history over the preceding ${\sim}6$\,s to estimate the viewing direction, and the KF output smooths the trajectory without replacing the GPS-derived prior.

In \emph{Spotter}, no GPS signal is available. The prior $\boldsymbol{\pi}_t$ for frame $t$ is the previous filtered position $\mathbf{\hat{g}}_{t-1}^+$. Candidate retrieval is additionally constrained by the prebuilt GSV street graph: on straight segments, candidates are restricted to a forward/backward extension along the detected street; at intersections, a Dijkstra traversal explores all reachable nodes within the search radius. Because the prior is chained across frames, an incorrect prediction is absorbed into the KF state and can compound into subsequent searches, making outlier rejection critical.

Both modes enforce a final \textit{gating mechanism} that rejects any pose estimate whose spatial deviation from the predicted prior exceeds:
\begin{equation}
    \tau_t = \begin{cases}
        \sigma^{\text{GPS}}_t + v^{\text{gate}}_t\,\Delta t + m_{\text{PnP}} & \text{SpotterGPS,} \\
        v_{\max}\,\Delta t_{\text{succ}} + m_{\text{PnP}} & \text{Spotter,}
    \end{cases}
    \label{eq:jump_gate}
\end{equation}
where $ v^{\text{gate}}_t = \min(v^{\text{GPS}}_t \times 1.5,\, v_{\max})$\,m/s is the speed estimate with a safety margin, $\Delta t$ is the inter-frame interval, $\Delta t_{\text{succ}}$ is the time elapsed since the last successful localization, and $m_{\text{PnP}} = 22$\,m is a fixed margin accounting for single-view PnP residual error. A prediction is rejected if $\|\mathbf{\hat{g}}_t^+ - \mathbf{\hat{g}}_{t-1}^+\|_2 > \tau_t$. In Spotter, this gate is the primary safeguard against corrupting the chained KF state with a gross outlier.

All hyperparameters were determined empirically on a held-out validation subset of the Barcelona dataset and kept fixed across all reported experiments.

\section{EXPERIMENTS}

\subsection{Dataset Construction}
\label{sec:dataset}

To evaluate Spotter in dense urban settings, we introduce a novel pedestrian-centric dataset curated specifically to capture continuous building facades across diverse streets in the city of Barcelona with different levels of GPS signal accuracy. Standard visual localization benchmarks---such as automotive datasets (KITTI~\cite{geiger2013vision}, nuScenes~\cite{caesar2020nuscenes}) or micro-aerial vehicle datasets (EuRoC MAV~\cite{Burri25012016})---either prioritize road-level perspectives with distant, occluded structures or focus on indoor and aerial domains. To bridge this gap, our dataset was recorded from a pedestrian viewpoint using wearable smart glasses\footnote{\url{https://bielglasses.com}} equipped with a ZED stereo camera system, capturing HD images ($1280 \times 720$\,px) at 5\,FPS to yield dense, persistent coverage of facade geometry.

The dataset comprises $20$ sequences spanning three Barcelona neighborhoods---Gr\`acia ($7$ sequences), Guinardó ($6$), and Poblenou ($7$)---for a total of $15{,}616$ frames, roughly $52$ minutes of recording, and close to $4$\,km of cumulative pedestrian trajectory. The three zones differ in street layout and building density, providing varied facade appearance and a range of localization difficulties. Raw GPS positioning data was logged concurrently using an Android mobile application running during data acquisition. Table~\ref{tab:dataset} summarizes the per-zone statistics.

\begin{table}[t]
    \centering
    \caption{Per-zone statistics of the Barcelona wearable dataset.}
    \label{tab:dataset}
    \begin{tabular}{lccccc}
        \toprule
        \textbf{Zone} & \textbf{Seq.} & \textbf{Frames} & \textbf{Dur.} & \textbf{Length} & \textbf{GPS err.} \\
             &      &        & \textbf{(min)}& \textbf{(m)}       & \textbf{(mean, m)}\\
        \midrule
        Gr\`acia  & 7 & 7041 & 23.6 & 1825 & 13.0 \\
        Guinard\'o  & 6 & 4323 & 13.9 & 1061 &  3.8 \\
        Poblenou  & 7 & 4252 & 14.4 & 1093 &  3.9 \\
        \midrule
        Total     & 20 & 15616 & 51.9 & 3979 & 7.0 \\
        \bottomrule
    \end{tabular}
\end{table}

A key feature of our dataset is its heterogeneous GPS quality across urban environments. While Guinard\'o and Poblenou maintain strong reception (mean horizontal errors of $3.8$\,m and $3.9$\,m), Gr\`acia exhibits severe signal degradation due to its narrow streets and high building density. Errors within Gr\`acia range from $6\text{--}7$\,m in Sequences 01–02 up to $14\text{--}20$\,m in Sequences 16–20 under heavy multipath conditions. Overall, Gr\`acia achieves a mean error of $13.0$\,m, contributing to a dataset-wide mean of $7.0$\,m—a trend consistent with established GPS error metrics in urban canyons~\cite{wen2021graphgnss}.

Ground-truth trajectories are available for all sequences. Each trajectory was manually annotated by cross-referencing the onboard GPS track with reference map data, providing complete ground-truth coverage across all 20 sequences and enabling consistent metric evaluation throughout every evaluation zone.

\subsection{Evaluation Protocol and Metrics}
\label{sec:protocol}

All methods are evaluated against the manually annotated ground-truth trajectories using the following metrics:

\paragraph{Absolute Position Error (APE).} The APE at frame $t$ is the Euclidean distance between the estimated global position $\mathbf{\hat{g}}_t^+ \in \mathbb{R}^2$ and the ground-truth position $\mathbf{{g}}_t^\text{GT} \in \mathbb{R}^2$ in the local UTM frame:
\begin{equation}
    \mathrm{APE}(t) = \bigl\|\mathbf{\hat{g}}_t^+ - \mathbf{{g}}_t^\text{GT}\bigr\|_2 .
    \label{eq:ape}
\end{equation}

\paragraph{Recall at distance thresholds.} To characterize the tail of the error distribution, we report recall at $3$, $5$, and $10$\,m (R@3\,m, R@5\,m, R@10\,m), defined as the fraction of frames for which the estimated position falls within the corresponding radius of the ground truth.

\paragraph{Efficiency.} We report the processing rate in frames per second (FPS), measured on the same hardware used for all experiments.

\begin{figure*}[ht!]
    \centering
        \includegraphics[width=0.245\textwidth]{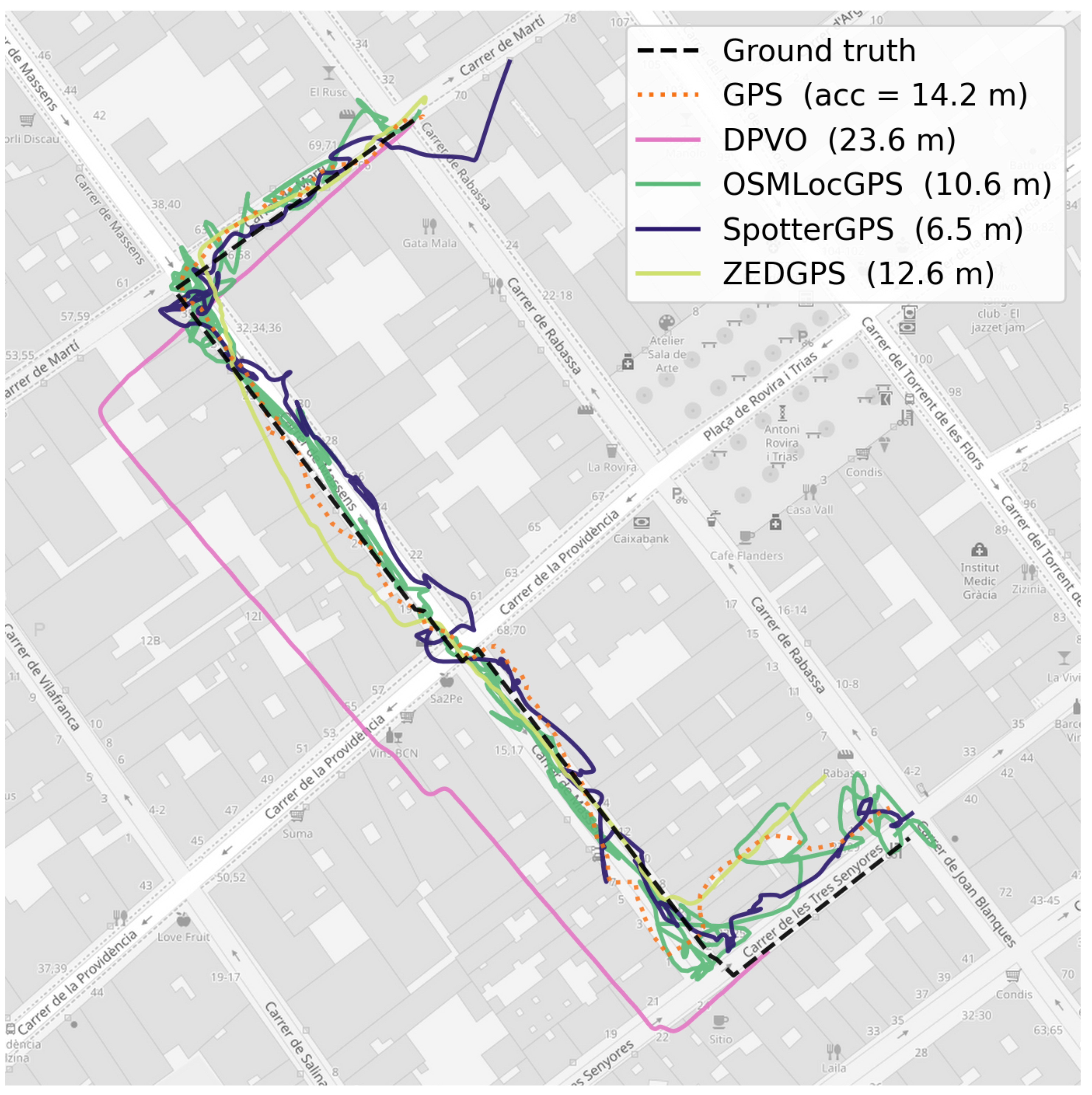}\hfill
        \includegraphics[width=0.245\textwidth]{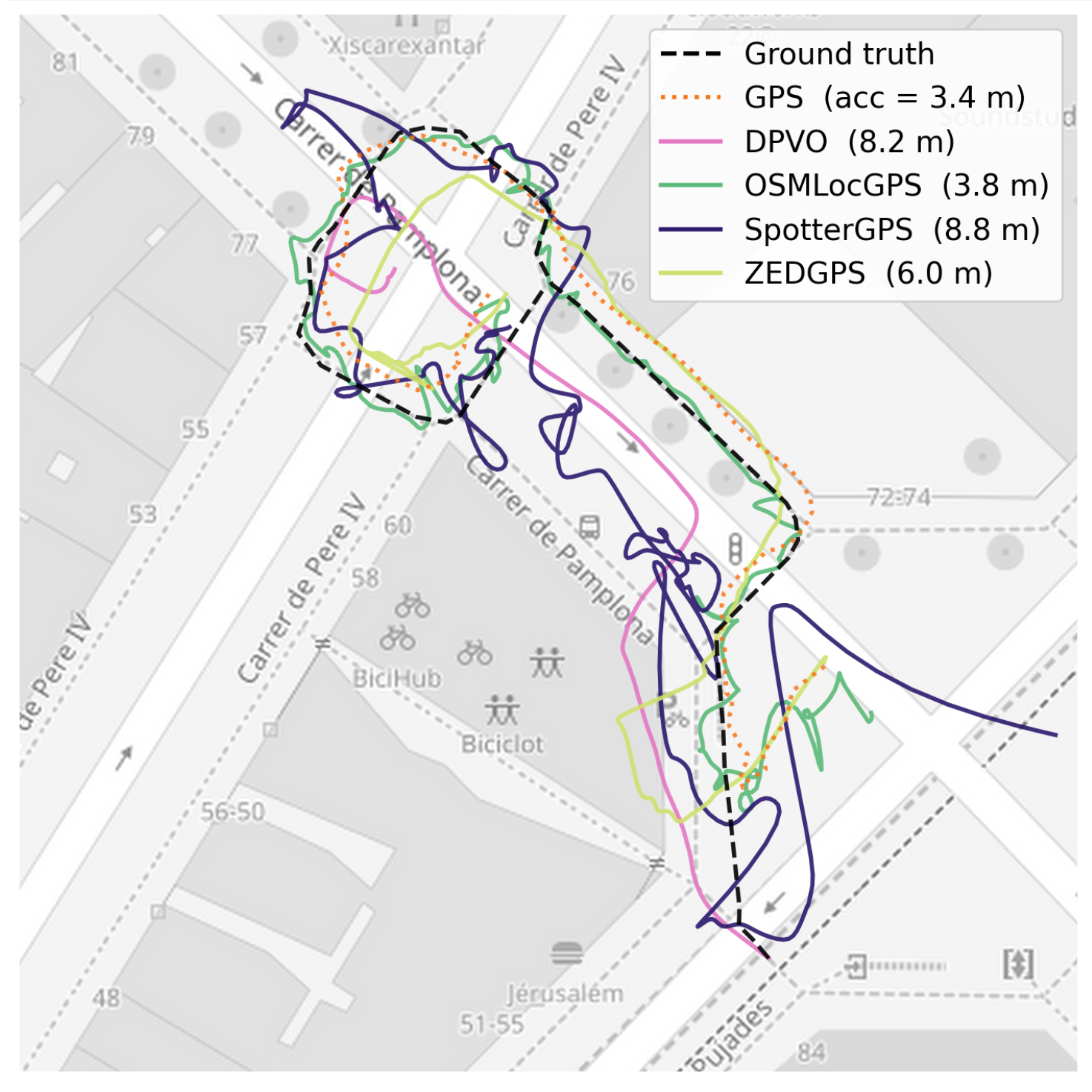}\hfill
        \includegraphics[width=0.245\textwidth]{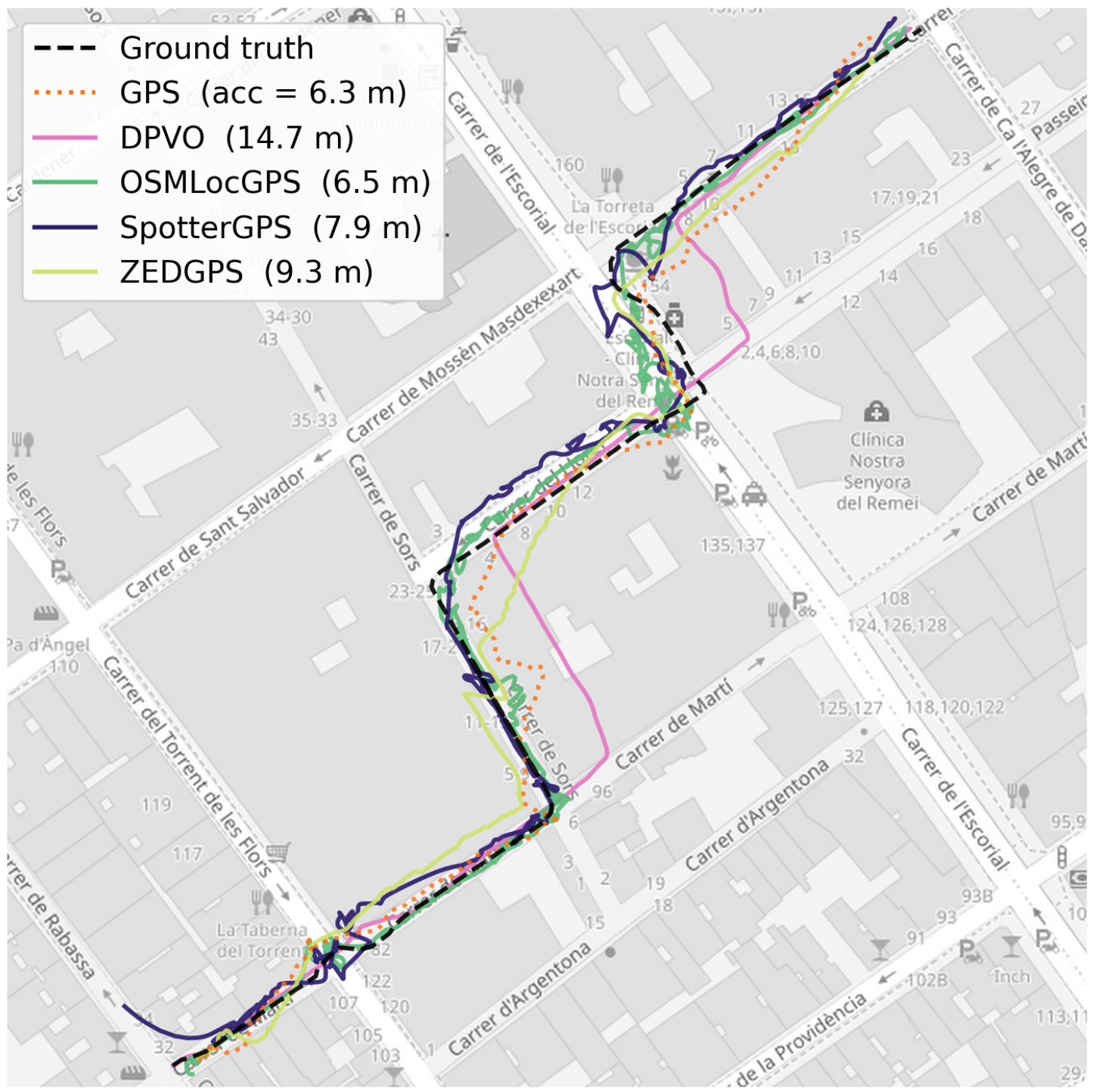}\hfill
        \includegraphics[width=0.245\textwidth]{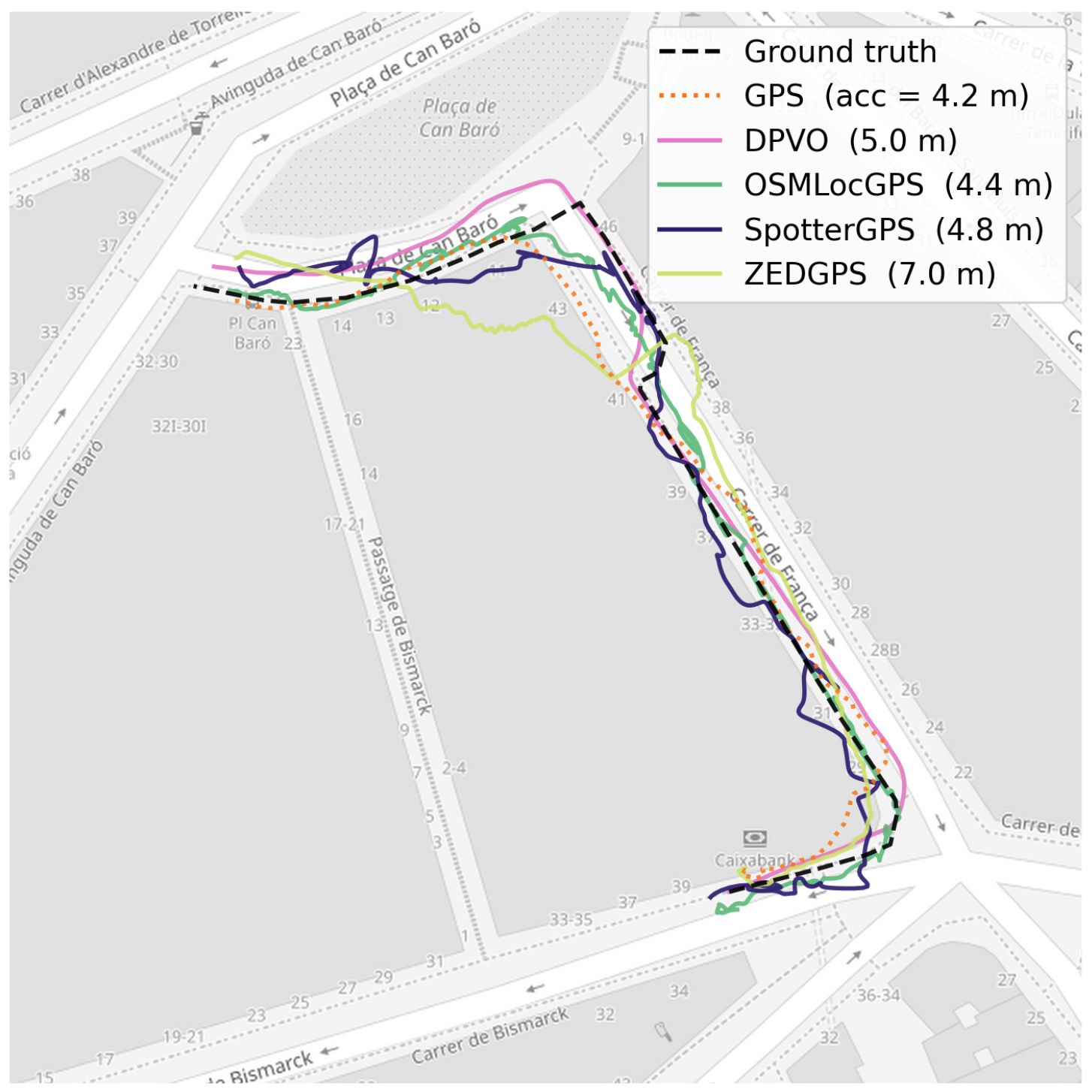}%
    \caption{Qualitative trajectory comparison across four sequences from different Barcelona neighborhoods under real GPS conditions. We show one representative method per paradigm. SpotterGPS (dark blue) consistently tracks the ground truth (black dashed), while DPVO (pink) accumulates drift and ZEDGPS (yellow) is limited by the same GPS multipath affecting each sequence. OSMLocGPS's performance (green) closely reflects the GPS accuracy of each sequence --- performing well where reception is good and degrading where multipath is severe.}
        \label{fig:traj_overall}
\end{figure*}

\subsection{Baselines}
\label{sec:baselines}

We benchmark Spotter against representative baselines across each localization paradigm using their default parameter configurations, executing all experiments on a workstation equipped with a single NVIDIA GeForce RTX 4090 GPU. Because several baselines perform single-frame global localization rather than continuous trajectory tracking, we additionally evaluate variants augmented with a GPS spatial prior (denoted as \textit{MethodGPS}). We group all evaluated methods into five distinct categories:

\begin{itemize}
\item \textbf{Proposed Method:} Spotter
\item \textbf{Sensor-Based:} ZED~\cite{zed_sdk_tracking}
\item \textbf{VO/SLAM:} TartanVO~\cite{tartan}, DPVO~\cite{dpvo}, MASt3R-SLAM~\cite{murai2024_mast3rslam}
\item \textbf{Map-Based:} OrienterNet~\cite{sarlin2023orienternet}, OSMLoc~\cite{liao2024osmloc}
\item \textbf{Retrieval-Based:} AnyLoc~\cite{keetha2023anyloc}
\end{itemize}

\subsection{Overall Results}
\label{sec:results}

Table~\ref{tab:overall} reports overall results across all 20 sequences. Among GPS-aided methods, OSMLocGPS achieves the lowest mean APE ($6.89$\,m) and the highest recall at all thresholds (R@3m $= 23.9\%$, R@5m $= 46.3\%$, R@10m $= 80.9\%$), benefiting from the reliable GPS signal that dominates the dataset. SpotterGPS achieves a mean APE of $8.13$\,m with R@10m of $71.6\%$, outperforming OrienterNetGPS ($10.00$\,m) while running at $45.5$ FPS---more than three times faster than the map-based baselines. However, as we show in Section~\ref{sec:gps_ablation}, the OSMLoc advantage collapses rapidly as GPS quality degrades, while SpotterGPS degrades far more gracefully.

Compared to the sensor-based and standalone VO/SLAM approaches, our method consistently achieves superior localization accuracy while maintaining a high processing frame rate—outperforming even DPVO in computational efficiency. As illustrated in Figure~\ref{fig:traj_overall}, methods that lack mechanisms for mitigating trajectory drift—such as global GPS priors or persistent visual landmarks—are susceptible to the unbounded accumulation of localization error over time. Note that for pure odometry and SLAM baselines, trajectories are scale-aligned to the ground truth via a 2D Similarity $\mathrm{Sim}(2)$ transformation using Umeyama alignment prior to computing the APE to ensure a fair comparison.

\begin{table}[t]
    \centering
    \caption{Overall localization results (20 sequences, 15{,}616 frames). VO/SLAM methods are Sim(2)-aligned before APE is computed. \textbf{Bold}: best per group.}
    \label{tab:overall}
    \resizebox{\columnwidth}{!}{%
    \begin{tabular}{lccccc}
        \toprule
        \textbf{Method} & \textbf{APE↓(m)} & \textbf{R@3m↑} & \textbf{R@5m↑} & \textbf{R@10m↑} & \textbf{FPS↑} \\
        \midrule
        \multicolumn{6}{l}{\textit{\textbf{Ours}}} \\
        SpotterGPS         & \textbf{8.13} & \textbf{15.7} & \textbf{32.0} & \textbf{71.6} & \textbf{45.5} \\
        Spotter            & 12.17 & 18.1 & 35.5 & 64.4 & 26.0 \\
        \midrule
        \multicolumn{6}{l}{\textit{Sensor-based}} \\
        ZEDGPS           & \textbf{12.36} &  3.1 &  9.9 & 48.3 & 5.0 \\
        ZED              & 15.90 & \textbf{15.7} & \textbf{26.5} & \textbf{51.5} & 5.0 \\
        \midrule
        \multicolumn{6}{l}{\textit{VO/SLAM}} \\
        TartanVO           & 38.97 &  6.9 & 13.6 & 34.3 & 28.1 \\
        DPVO               &  \textbf{9.78} & \textbf{18.9} & \textbf{37.7} & \textbf{65.9} & \textbf{43.8} \\
        MASt3R-SLAM        & 30.75 &  8.1 & 14.8 & 32.9 & 0.8 \\
        \midrule
         \multicolumn{6}{l}{\textit{Map-based}} \\
        OrienterNetGPS     & 10.00 &  7.2 & 17.9 & 58.9 & \textbf{13.0} \\
        OrienterNet        & 88.88 &  7.2 & 11.2 & 16.1 & 10.0 \\
        OSMLocGPS          &  \textbf{6.89} & \textbf{23.9} & \textbf{46.3} & \textbf{80.9} & 12.7 \\
        OSMLoc             & 45.98 & 21.7 & 29.0 & 36.1 & 12.7 \\
        \midrule
        \multicolumn{6}{l}{\textit{Retrieval-based}} \\
        AnyLoc             & 121.65 & 6.7 & 14.7 & 30.4 & \textbf{2.3} \\
        AnyLocGPS          & \textbf{11.71} & \textbf{8.4} & \textbf{21.5} & \textbf{56.1} & 2.0 \\
        \bottomrule
    \end{tabular}}
\end{table}

Among map-based methods, we observe a substantial disparity in performance depending on whether an accurate GPS prior is available. Since these approaches are primarily designed for discrete map-based localization rather than continuous trajectory estimation, the absence of a reliable GPS prior requires the search to cover substantially larger map regions. This increases the number of candidate pose hypotheses and introduces greater visual ambiguity, which can lead to severe mislocalizations. Moreover, evaluating visual-spatial correspondences across such extensive candidate regions incurs considerable computational overhead, substantially reducing inference speed.

\begin{figure*}[ht!]
    \centering
        \includegraphics[width=0.245\textwidth]{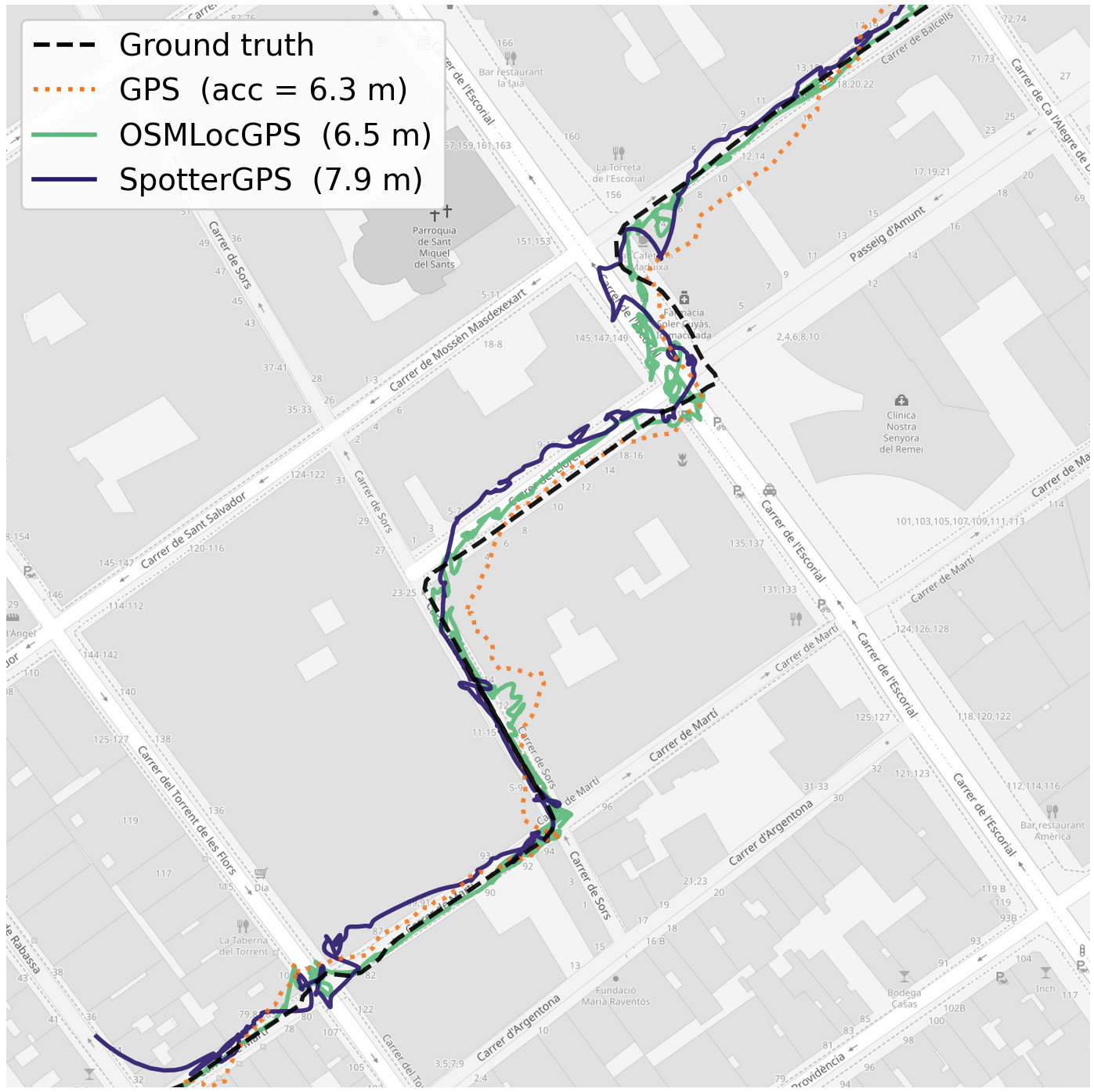}\hfill
        \includegraphics[width=0.245\textwidth]{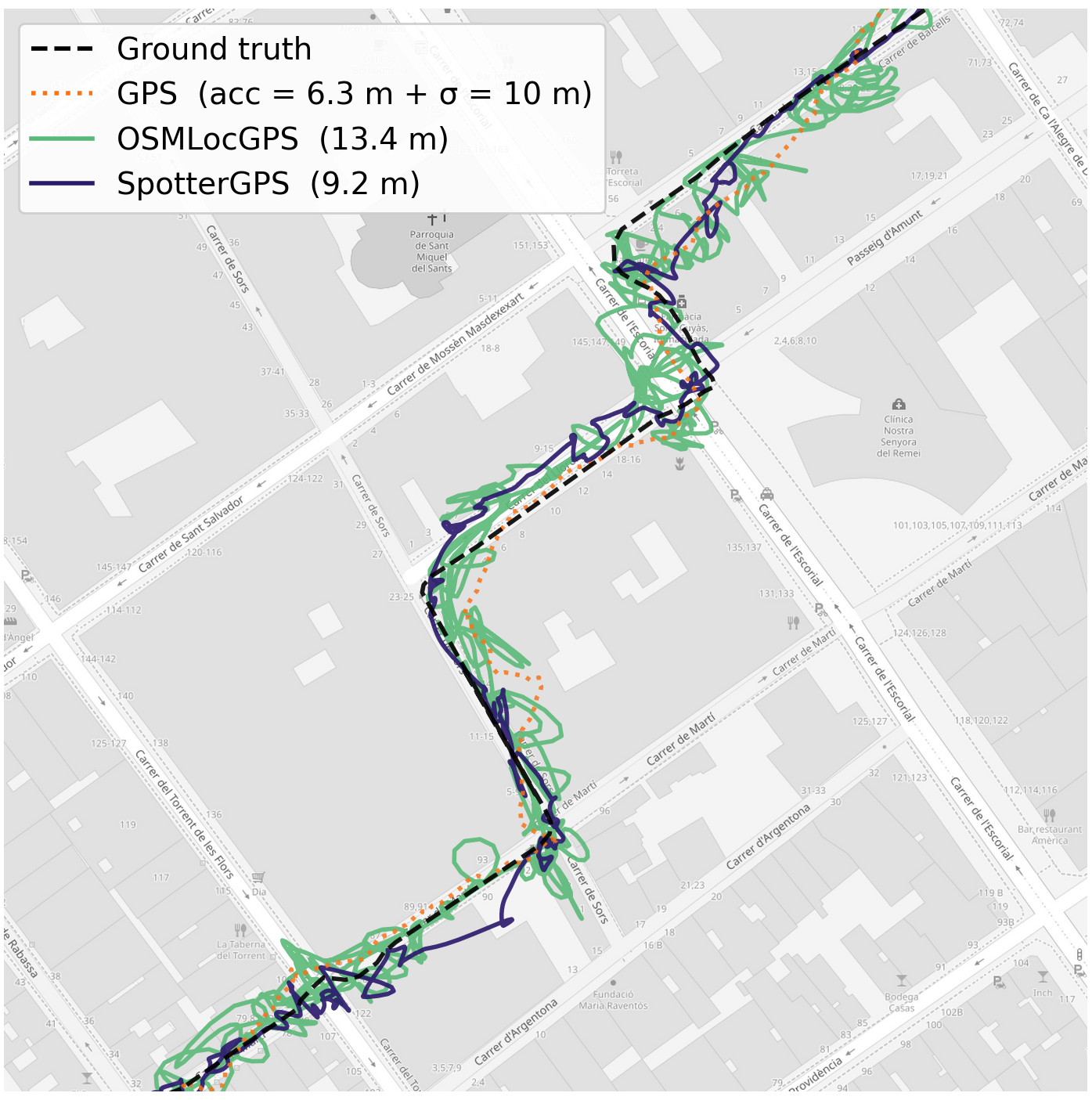}\hfill
        \includegraphics[width=0.245\textwidth]{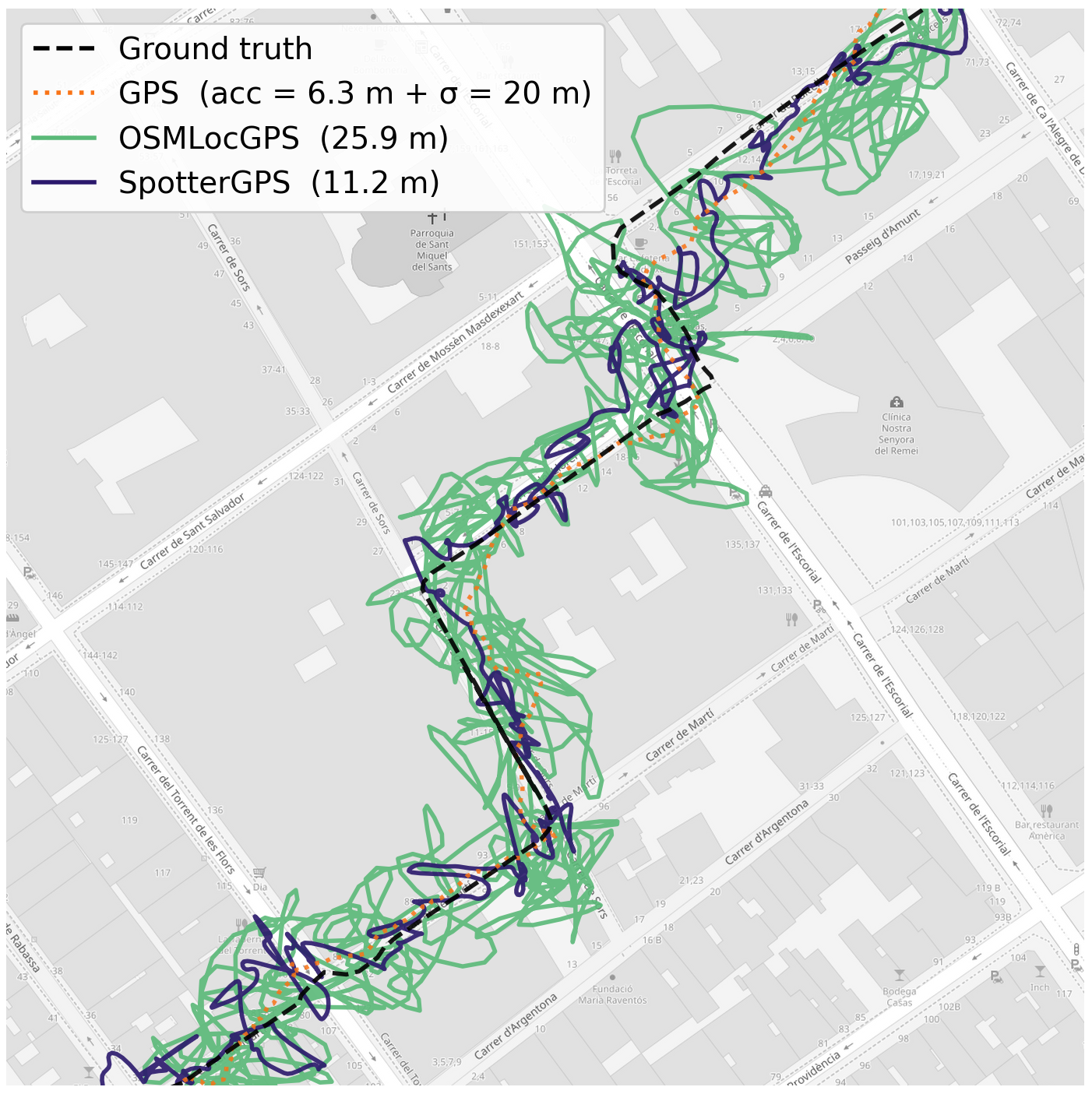}\hfill
        \includegraphics[width=0.245\textwidth]{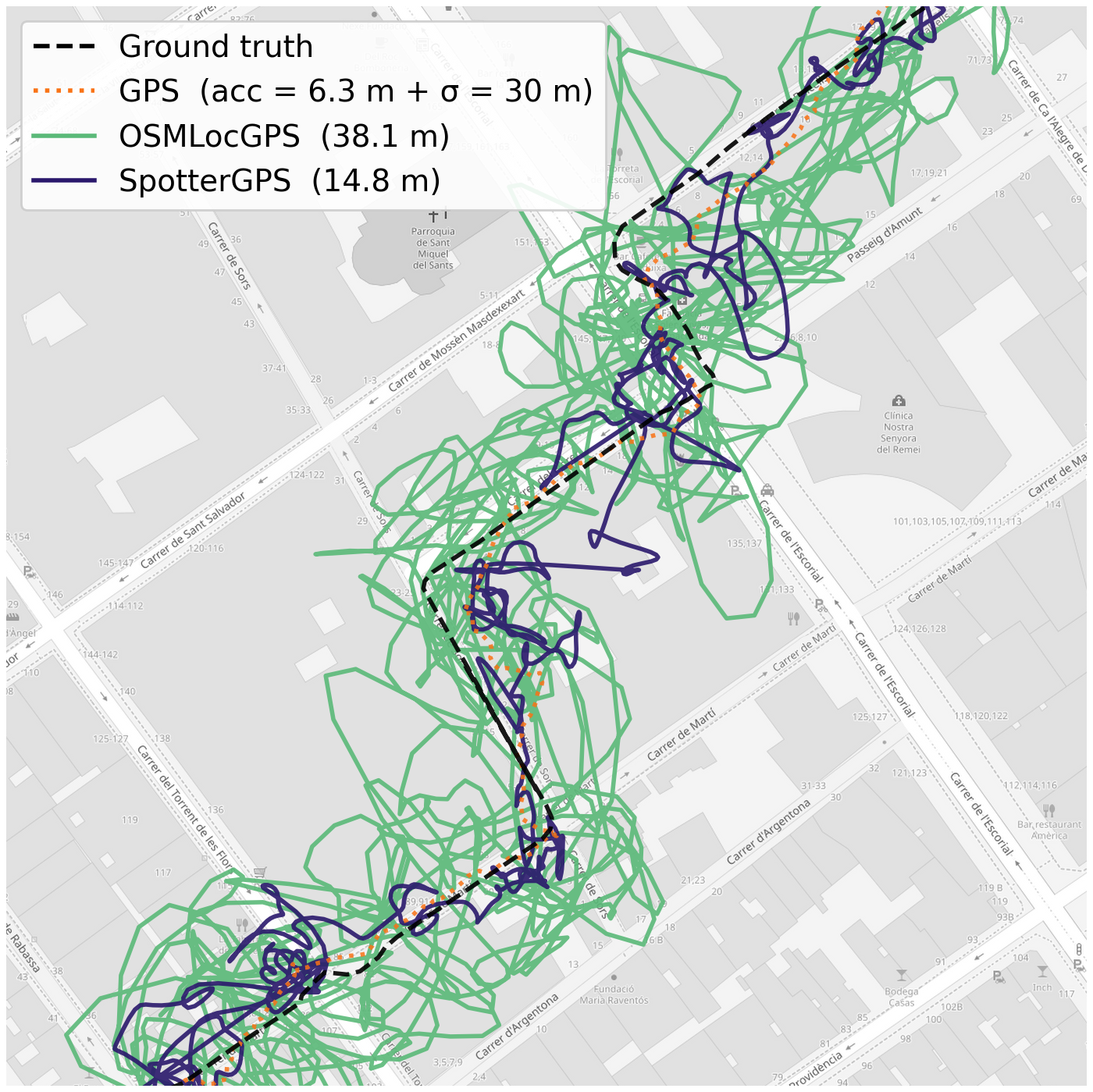}%
    \caption{GPS noise ablation on a single Gràcia sequence with injected noise $\sigma \in \{0, 10, 20, 30\}$\,m. At $\sigma = 0$\,m, OSMLocGPS (green) achieves competitive accuracy ($6.5$\,m vs.\ $7.6$\,m for SpotterGPS); as noise increases, OSMLocGPS degrades rapidly with erratic jumps and off-road detours, while SpotterGPS (dark blue) remains broadly coherent.}
    \label{fig:trajectories} 
\end{figure*}

\begin{figure}[ht!]
    \centering
    \includegraphics[width=\columnwidth]{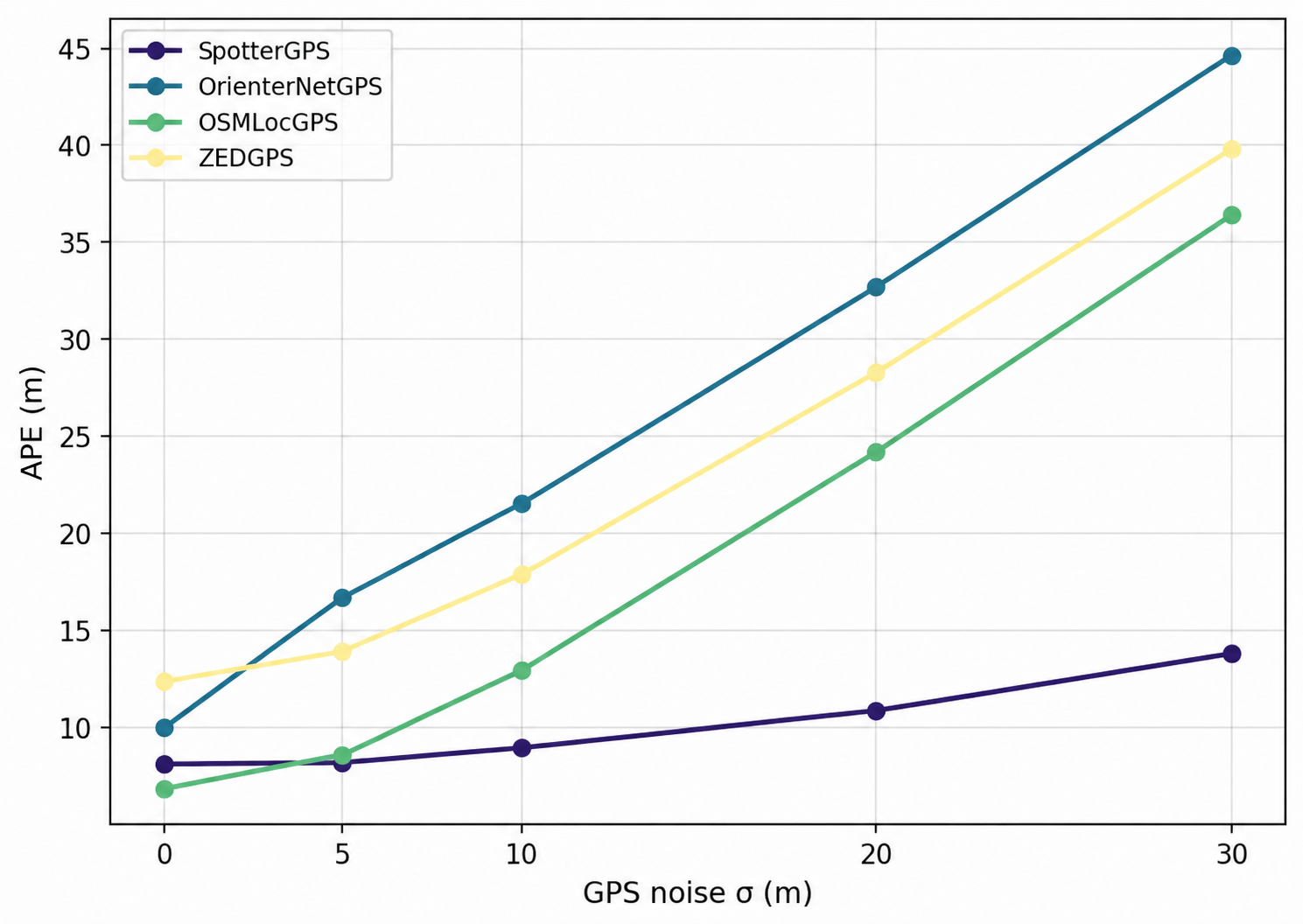}
    \caption{Mean APE vs.\ injected GPS noise $\sigma$ for all GPS-dependent methods. SpotterGPS (purple) remains nearly flat up to $\sigma=10$\,m and degrades slowly beyond, while OrienterNetGPS (blue), ZEDGPS (yellow), and OSMLocGPS (green) rise steeply, confirming Spotter's robustness to GPS degradation.}
    \label{fig:gps_ablation}
\end{figure}

The retrieval-based baseline AnyLoc reaches $121.65$\,m APE, by far the largest error in the table. Because it reports the coordinate of the nearest retrieved panorama and applies no spatial prior, a single ambiguous retrieval can return a viewpoint elsewhere in the city; the repetitive facades of Barcelona's regular street grid make such confusions frequent. Restricting retrieval to the GPS prior (AnyLocGPS) removes these gross confusions and cuts the error to $11.71$\,m---a tenfold reduction from a spatial prior alone, with no change to the descriptors or database. Yet recall at tight thresholds stays low ($8.4\%$ at $3$\,m), because retrieval can at best return the nearest reference viewpoint, sampled at $30$\,m intervals: even with a correct prior it cannot place the camera more finely than the database sampling allows. AnyLocGPS therefore remains well behind SpotterGPS ($8.13$\,m APE, $32.0\%$ R@5m vs.\ $21.5\%$) under the same prior, showing that the explicit 2D--3D correspondences and PnP solver---not the retrieval or the prior---are what deliver metric pedestrian localization.

Figure~\ref{fig:traj_overall} provides a graphical comparison of the different methods, together with the ground-truth trajectories and GPS measurements, for four sequences from the dataset. Overall, the qualitative results are consistent with the conclusions drawn from the numerical evaluation.

\subsection{Robustness to GPS Noise}
\label{sec:gps_ablation}

The preceding results reveal a strong correlation between GPS signal quality and overall localization accuracy across all evaluated baselines. For instance, OSMLocGPS performs exceptionally well when provided with an accurate initial prior, as the search region remains tightly constrained near the true location. However, in regions characterized by repetitive urban structures, increasing GPS noise expands the spatial search radius. This introduces perceptual ambiguities and multiple candidate pose hypotheses, ultimately leading to higher localization errors. These findings motivate a systematic evaluation of method robustness under varying noise regimes, demonstrating the advantage of using persistent building facades as landmark anchors to maintain localization precision.

To isolate the effect of GPS quality from other factors such as scene complexity or sequence length, we conduct a controlled ablation: we inject zero-mean Gaussian noise with standard deviation $\sigma \in \{0, 5, 10, 20, 30\}$\,m into the GPS readings of all sequences and re-evaluate each GPS-dependent method under these degraded priors. This transforms GPS noise from a scene-dependent confound into a controlled experimental variable, letting us directly compare the robustness of different GPS-reliant approaches. Note that for SpotterGPS, larger GPS noise also increases the candidate search radius, which directly raises the number of candidates to evaluate and reduces throughput; this effect is visible in the FPS column of Table~\ref{tab:gps_ablation}.

\begin{table}[t!]
    \caption{GPS noise ablation. APE and recall for GPS-dependent methods at five noise levels. \textbf{Bold}: best at each level.}
    \label{tab:gps_ablation}
    \resizebox{\columnwidth}{!}{%
    \begin{tabular}{llccccc}
        \toprule
        $\sigma$\,\textbf{(m)} & \textbf{Method} & \textbf{APE↓(m)} & \textbf{R@3m↑} & \textbf{R@5m↑} & \textbf{R@10m↑} & \textbf{FPS↑} \\
        \midrule
        \multirow{4}{*}{0}
          & SpotterGPS     & 8.13  & 15.7          & 32.0          & 71.6          & \textbf{45.5} \\
          & OrienterNetGPS & 10.00 & 7.2           & 17.9          & 58.9          & 13.0 \\
          & OSMLocGPS      & \textbf{6.89}  & \textbf{23.9} & \textbf{46.3} & \textbf{80.9} & 12.7 \\
          & ZEDGPS        & 12.36 & 3.1           & 9.9           & 48.3          & 5.0  \\
        \midrule
        \multirow{4}{*}{5}
          & SpotterGPS     & \textbf{8.21}  & 14.6 & 31.8 & \textbf{70.3} & \textbf{46.3}  \\
          & OrienterNetGPS & 16.70 & 3.6           & 8.1           & 28.1          & 13.0  \\
          & OSMLocGPS      & 8.62  & \textbf{18.4}          & \textbf{35.2}          & 69.8          & 12.7  \\
          & ZEDGPS        & 13.94 & 4.2           & 11.1          & 37.6          & 5.0  \\
        \midrule
        \multirow{4}{*}{10}
          & SpotterGPS     & \textbf{8.97}  & \textbf{13.5} & \textbf{28.1} & \textbf{65.9} & \textbf{42.3}  \\
          & OrienterNetGPS & 21.52 & 1.6           & 4.5           & 17.8          & 12.3  \\
          & OSMLocGPS      & 12.94 & 11.3          & 22.0          & 48.8          & 13.0  \\
          & ZEDGPS        & 17.87 & 2.4           & 6.7           & 24.5          & 5.0  \\
        \midrule
        \multirow{4}{*}{20}
          & SpotterGPS     & \textbf{10.87} & \textbf{9.0}  & \textbf{20.7} & \textbf{53.7} & \textbf{38.0}  \\
          & OrienterNetGPS & 32.72 & 0.4           & 1.9           & 7.2           & 12.1  \\
          & OSMLocGPS      & 24.20 & 5.3           & 10.8          & 25.6          & 12.7  \\
          & ZEDGPS        & 28.30 & 0.9           & 2.6           & 10.0          & 5.0  \\
        \midrule
        \multirow{4}{*}{30}
          & SpotterGPS     & \textbf{13.80} & \textbf{5.2}  & \textbf{13.3} & \textbf{42.0} & \textbf{36.1} \\
          & OrienterNetGPS & 44.68 & 0.2           & 1.0           & 3.7           & 12.0  \\
          & OSMLocGPS      & 36.43 & 3.4           & 6.7           & 16.1          & 12.8  \\
          & ZEDGPS        & 39.84 & 0.4           & 1.4           & 5.2           & 5.0  \\
        \bottomrule
    \end{tabular}}
\end{table}

Figure~\ref{fig:trajectories} shows qualitative trajectory comparisons on a sequence from Gràcia; Figure~\ref{fig:gps_ablation} plots performance curves for all GPS-dependent methods as a function of injected noise $\sigma$; and Table~\ref{tab:gps_ablation} provides the full numerical breakdown at each noise level.

At $\sigma = 0$\,m (no added noise), OSMLocGPS leads in recall (R@3m $= 23.9\%$, R@5m $= 46.3\%$) and APE ($6.89$\,m), while SpotterGPS trails slightly at $8.13$\,m. As noise increases to $\sigma = 5$\,m, SpotterGPS already becomes the best overall (APE $8.21$\,m, R@3m $14.6\%$, R@10m $70.3\%$), while OSMLocGPS drops to $8.62$\,m APE and R@10m falls from $80.9\%$ to $69.8\%$. By $\sigma = 10$\,m, the gap widens decisively: SpotterGPS holds at $8.97$\,m APE and R@10m $65.9\%$, while OSMLoc degrades to $12.94$\,m and OrienterNetGPS to $21.52$\,m. At $\sigma = 30$\,m---corresponding to the worst multipath conditions observed in Gr\`acia---SpotterGPS achieves $13.80$\,m APE and R@10m of $42.0\%$, while OrienterNet ($44.68$\,m), OSMLoc ($36.43$\,m), and ZEDGPS ($39.84$\,m) exhibit near-complete localization failure.

This pattern reflects a fundamental architectural difference: OSMLocGPS and OrienterNetGPS use the GPS coordinate to select the OSM tile against which the query is matched, so a large GPS error directly corrupts the input to the matching stage. Spotter instead uses GPS only to define a candidate search radius; the final pose is determined entirely by visual feature matching and PnP, so the system degrades gracefully and continues to localize correctly even when the prior is tens of meters off. When GPS is perfect or near-perfect, OSMLoc provides slightly higher recall at tight thresholds; but for any realistic urban deployment---where GPS errors of $5$--$20$\,m are common due to multipath, tall buildings, and signal blockage---SpotterGPS is consistently preferable. GPS is used as a coarse geographic hint, not as a trusted measurement, and visual evidence always takes precedence in determining the final pose.

\subsection{GPS Denied Areas}
As illustrated in Table~\ref{tab:overall}, Spotter achieves a mean APE of $12.17$\,m with R@10m of $64.4\%$, outperforming both sensor-based (ZEDGPS: $12.36$\,m; ZED: $15.90$\,m) and the remaining GPS-free localization methods, despite using no satellite signals. DPVO remains the only GPS-free baseline with a lower APE ($9.78$\,m), though as noted above its result requires a global $\mathrm{Sim}(2)$ alignment to ground truth and is therefore not achievable in an online setting.

\begin{figure}[t!]
    \centering
    \includegraphics[width=0.24\textwidth]{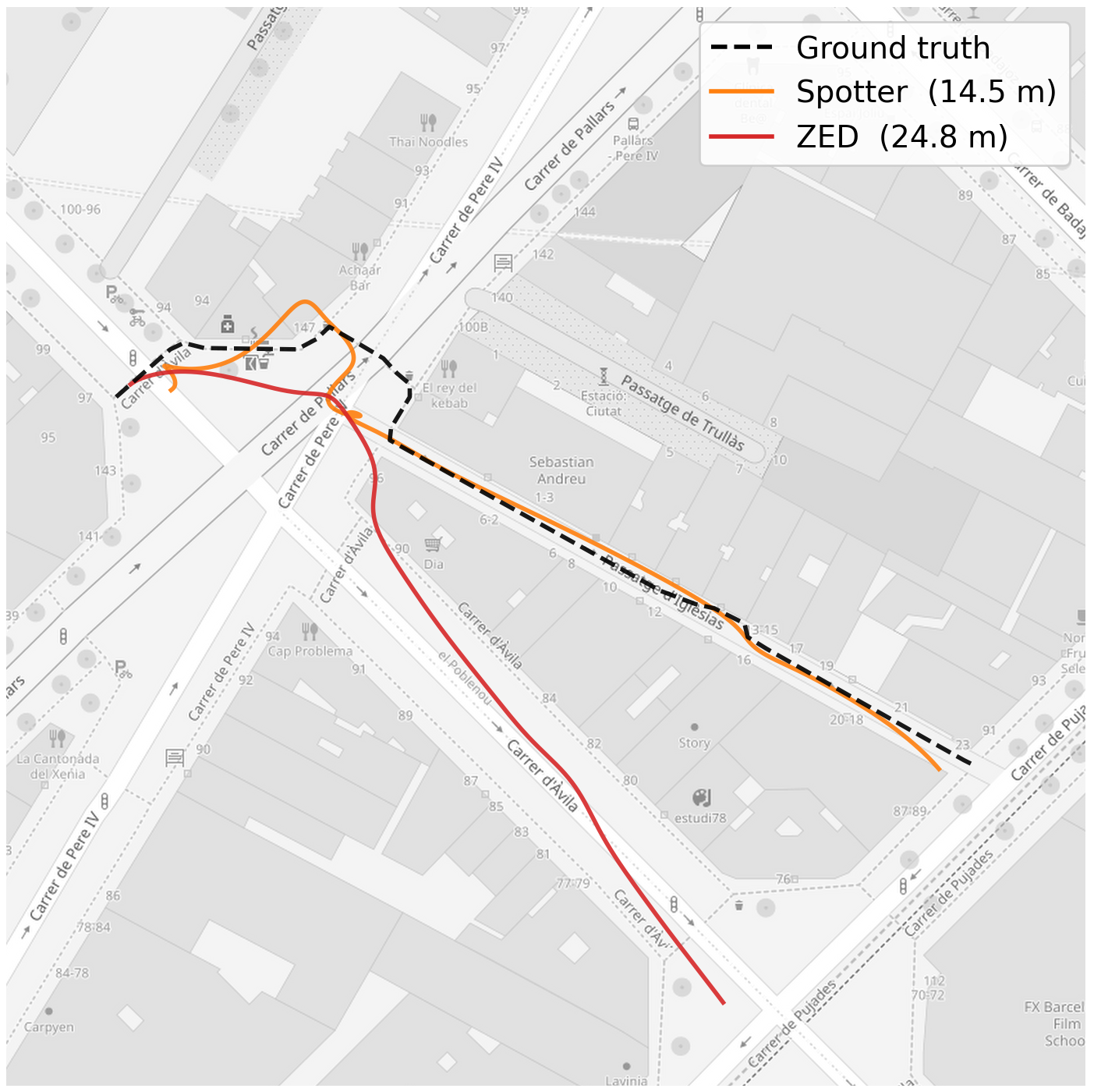}\hfill
    \includegraphics[width=0.24\textwidth]{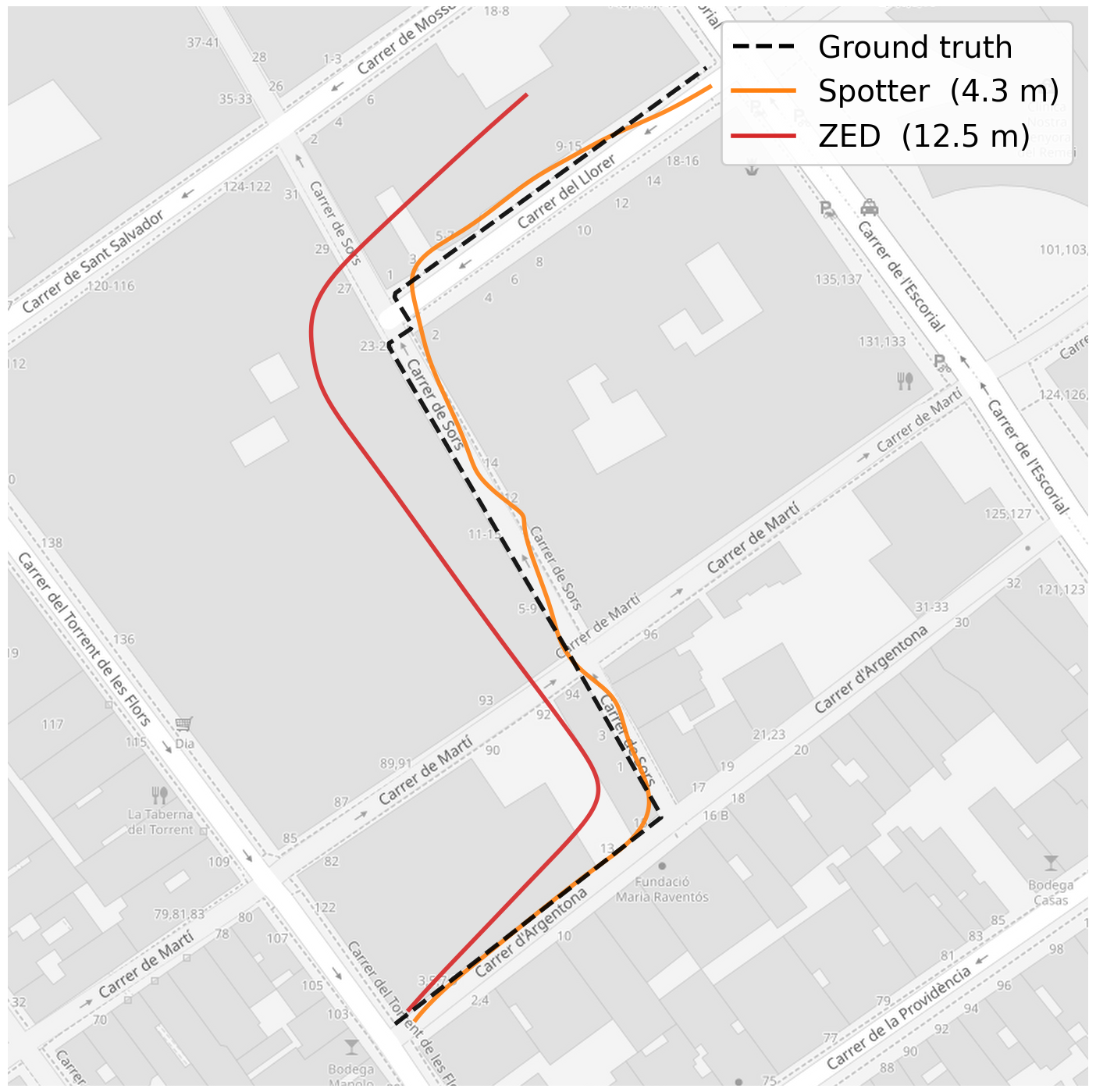}
    \caption{Qualitative trajectory comparison between Spotter (orange) and ZED (red) against ground truth (black dashed) on two representative sequences, both operating without any GPS signal. \emph{Left}: ZED drifts onto an adjacent street after a turn, while Spotter recovers and continues to track the correct path. \emph{Right}: both methods follow the ground-truth trajectory closely, with Spotter achieving lower error along the full sequence.}
    \label{fig:traj_gpsminus}
\end{figure}

Therefore, we visually compare Spotter with the best-performing GPS-free method, the sensor-based ZED method, to examine the behavior of both approaches in GPS-denied environments. As illustrated in Figure~\ref{fig:traj_gpsminus}, when ZED loses track during the early stages of the trajectory, Spotter remains close to the ground-truth path. Moreover, although Spotter exhibits small amounts of drift in some portions of the trajectory, it is able to subsequently recover and return to the correct path, demonstrating its robustness in the absence of GPS.


\section{CONCLUSIONS}

We presented Spotter, a lightweight visual localization framework designed for urban navigation on resource-constrained wearable platforms. Spotter estimates global camera poses from street-level images by matching query frames against a compact database of geo-referenced Google Street View (GSV) panoramas, bypassing the need for computationally heavy 3D scene reconstructions. A central pillar of its design is the strict decoupling of the GPS signal from visual matching: GPS serves solely as a coarse spatial filter, leaving the metric pose to be determined entirely via 2D--3D correspondences solved through PnP/RANSAC. Consequently, Spotter maintains high accuracy in complex urban environments where traditional tile-based or GPS-dependent retrieval pipelines break down.

Comprehensive evaluation on a wearable dataset spanning three distinct neighborhoods in Barcelona validates our design choices. Under reliable GPS coverage, SpotterGPS matches the accuracy of state-of-the-art map-based methods while delivering significantly lower computational overhead. Controlled ablations under identical spatial priors demonstrate that these performance gains stem directly from our robust pose-estimation pipeline rather than the retrieval prior alone. Furthermore, as GPS signals degrade, Spotter exhibits graceful failure modes, while its GPS-free variant (Spotter) continues to localize effectively without satellite assistance—outperforming the sensor-based pipeline and directly validating permanent building facade features as a sufficient geometric constraint for global metric localization.

Future research will focus on three key enhancements. First, because urban building facades are predominantly planar, the resulting 2D--3D correspondences can become geometrically ill-conditioned; expanding the database to include non-planar structural landmarks such as ground markings and road signage will improve pose estimation stability. Second, explicitly detecting and filtering dynamic occluders, such as pedestrians and parked vehicles, will bolster robustness in densely populated environments. Finally, benchmarking across multi-city datasets, seasonal variations, and extreme illumination changes will establish the global scalability and long-term generalization of our approach.

\section*{Acknowledgments}
\noindent Project PCPP2021-008760 funded by MCIU/ AEI /10.13039/501100011033 and by the "European Union NextGenerationEU/PRTR"

\bibliographystyle{IEEEtran}  
\bibliography{references}

\begin{IEEEbiography}[{\includegraphics[width=1in,height=1.25in,clip,keepaspectratio]{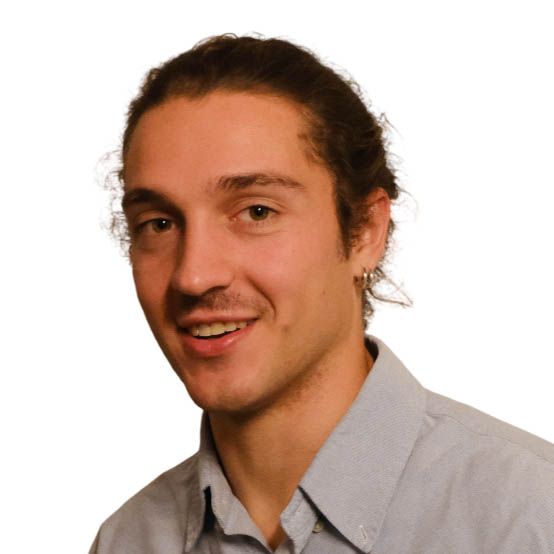}}]{Antoni Valls}
received the B.Sc. degree in theoretical physics from the 
Universitat de Barcelona, Spain, in 2022, and the 
M.Sc. degree in data science from the University of Padua,
Italy, in 2024. He has previously collaborated with the Institut d'Investigació en Intel·ligència Artificial (IIIA), CSIC, Barcelona, Spain, on natural language processing research. He is currently a Robotic Engineer with the Institut de Robòtica i Informàtica Industrial (IRI), CSIC-UPC, Barcelona, Spain, where he works on the SMARTGAZE~II project in collaboration with Biel Glasses, developing AI-powered assistive systems to enhance the mobility of people with low vision. His research interests include computer vision, vision-language models, assistive and pedestrian navigation, localization in dynamic environments, and deep learning.
\end{IEEEbiography}

\begin{IEEEbiography}[{\includegraphics[width=1in,height=1.25in,clip,keepaspectratio]{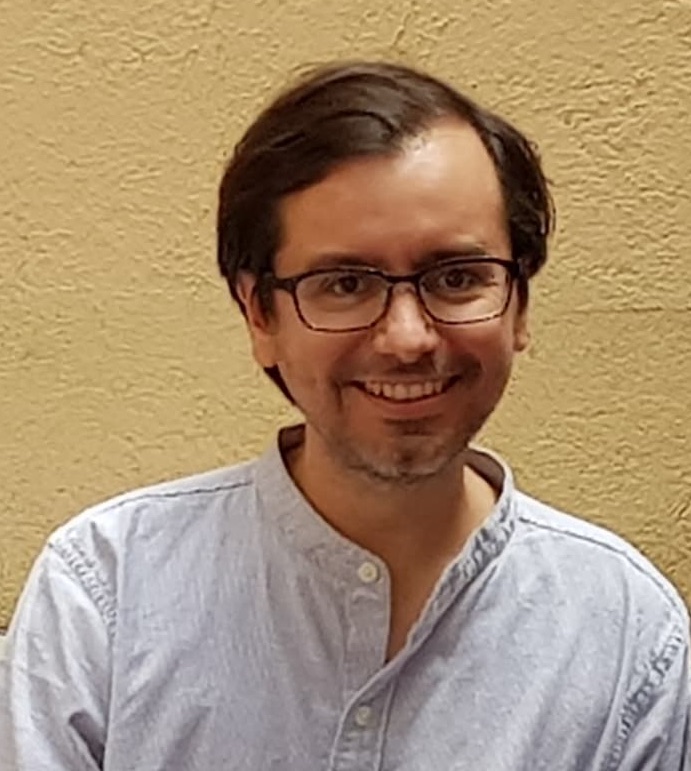}}]{Jordi Sanchez-Riera} is an associate researcher at the Spanish Scientific Research Council (CSIC) in Barcelona. He earned his B.S. in computer science and M.S. in electronic engineering from UPC, and a Ph.D. from the University of Grenoble. His research focuses on robotics, computer vision, and machine learning. He has published in top journals and conferences and received prestigious awards, including a Google Scholarship, Academia Sinica Post-Doc Fellowship, and Juan de la Cierva Incorporación grant.
\end{IEEEbiography}

\end{document}